\documentclass{article}

\usepackage[preprint]{neurips_2025}

\usepackage[utf8]{inputenc} 
\usepackage[T1]{fontenc}    
\usepackage{hyperref}       
\usepackage{url}            
\usepackage{booktabs}       
\usepackage{amsfonts}       
\usepackage{nicefrac}       
\usepackage{microtype}      
\usepackage{lipsum}		
\usepackage{graphicx}
\graphicspath{{Figures/}}
\usepackage{natbib}
\usepackage{doi}
\usepackage{amsmath}
\usepackage{amsthm}
\usepackage{dsfont}
\usepackage{bbm}
\usepackage{derivative}
\usepackage{epsfig}
\usepackage{subcaption}
\usepackage{multirow}
\usepackage{tablefootnote}
\usepackage{algorithm}
\usepackage{algpseudocode}
\usepackage[capitalize]{cleveref}
    \crefname{section}{Sec.}{Secs.}
    \Crefname{section}{Section}{Sections}
    \Crefname{table}{Table}{Tables}
    \crefname{table}{Tab.}{Tabs.}

\DeclareMathOperator*{\argmin}{arg\,min}

\newcommand{\tb}[1]{\textbf{#1}}

\newcommand{\ti}[1]{\textit{#1}}

\newcommand\norm[1]{\left\lVert#1\right\rVert}
\newcommand{\GP}{\tb{\ti{GP}}}

\newcommand*{\rom}[1]{\uppercase\expandafter{\romannumeral #1\relax}}

\title{Posterior shape models revisited: Improving 3D reconstructions from partial data using target specific models}

\date{}

\author{ \href{https://orcid.org/0000-0003-2418-5161}{\includegraphics[scale=0.06]{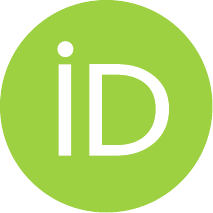}\hspace{1mm}Jonathan Aellen} \and \href{https://orcid.org/0000-0002-4304-4076}{\includegraphics[scale=0.06]{orcid.pdf}\hspace{1mm}Florian Burkhardt} \and {\hspace{1mm}Thomas Vetter} \and \href{https://orcid.org/0000-0002-9686-2195}{\includegraphics[scale=0.06]{orcid.pdf}\hspace{1mm}Marcel Lüthi} \\
Department of Mathematics and Computer Science\\
University of Basel \\ Switzerland, Basel 4001 \\ 
\texttt{jonathan.aellen@unibas.ch}
}

\hypersetup{
pdftitle={Posterior shape models revisited: Improving 3D reconstructions from partial data using target specific models},
pdfauthor={Jonathan Aellen, Florian Burkhardt, Thomas Vetter, Marcel Lüthi},
pdfkeywords={Uncertainty estimation, Partial shape reconstruction, Point distribution models},
}

\begin{document}
\maketitle

\begin{abstract}

In medical imaging, point distribution models are often used to reconstruct and complete partial shapes using a statistical model of the full shape. A commonly overlooked, but crucial factor in this reconstruction process, is the pose of the training data relative to the partial target shape. A difference in pose alignment of the training and target shape leads to biased solutions, particularly when observing small parts of a shape. In this paper, we demonstrate the importance of pose alignment for partial shape reconstructions and propose an efficient method to adjust an existing model to a specific target. Our method preserves the computational efficiency of linear models while significantly improving reconstruction accuracy and predicted variance. It exactly recovers the intended aligned model for translations, and provides a good approximation for small rotations, all without access to the original training data. Hence, existing shape models in reconstruction pipelines can be adapted by a simple preprocessing step, making our approach widely applicable in plug-and-play scenarios.
\end{abstract}


\section{Introduction}

Reconstruction of the full 3D shape from a partial observation is a common task in medical image analysis, computer vision, forensics and many other fields. In many of these areas high accuracy, an entire distribution over reconstructions, and exclusively plausible shapes are important. A well-established Bayesian approach to this problem is to use statistical shape models, where a prior distribution over plausible shapes is learned from example data. The reconstruction process is then formulated as posterior inference. However, the complexity of the posterior distribution, particularly when parameters are entangled, can make approximate Bayesian inference challenging. An often overlooked key aspect of this task is that the structure of the prior significantly affects the complexity of the inference.

If the prior is Gaussian, as is the case for point distribution models (PDMs), inference can be directly performed using the predictive posterior distribution. While more complex, non-linear priors might offer additional flexibility, linear models are already good approximators in many practical applications and crucially there is often not enough data to train an adequate non-linear model. Linear models provide accurate reconstructions, maintain computational efficiency and require significantly less training data. When the underlying shape variations are well captured by a Gaussian prior, the linear nature of PDMs allows direct inference without the need for costly optimization steps. The resulting distribution of possible shape reconstructions is often called a posterior shape model \citep{albrecht2013posterior}, which represents the updated belief after incorporating observed data.  However, if the training and target data are not consistently aligned in pose, the posterior shape model is strongly biased and systematically underestimates uncertainty. More critically, this also coincides with making the posterior during approximate inference unnecessarily complex. In this paper, we revisit posterior shape models and demonstrate that a simple modification of the prior results in a significantly more tractable inference process.

\begin{figure}%
    \centering
    \includegraphics[width=\linewidth]{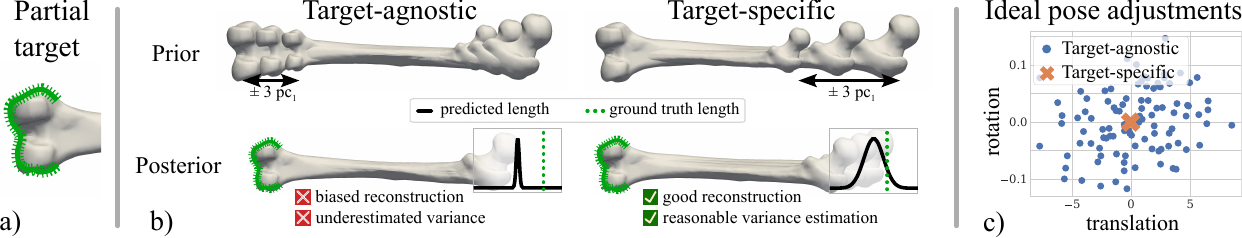}%
    \caption{We showcase the importance of the pose alignment of training data during partial target reconstruction using posterior shape models. a) The observed partial target, indicated by a green outline. b) Top: prior shape distribution. Bottom: posterior distributions over partial target reconstructions. All represented by overlaying the mean and $\pm$3 standard deviations of the first principal component. For partial targets, target-agnostic models have biased and overconfident reconstructions, while target-specific models have accurate reconstructions, demonstrated by the predicted and ground truth length. c) The plot shows the dependency between pose and shape, which plays an important role in reconstruction performance. The ideal pose adjustments to the target are shown for a number of samples from the shape prior. The shape of the target-specific model can be optimized independent of the pose, simplifying the problem greatly.}%
    \label{fig:fig1compact}%
\end{figure}

\Cref{fig:fig1compact} illustrates the impact of data alignment and resulting prior choice on reconstruction quality. Target-agnostic models, not aligned to the target, exhibit an entangled shape and pose space. Random samples from such a model require further pose optimization indicating why the inference is more difficult. In contrast, target-specific models, where the prior has been adapted to the target, have a clearly disentangled shape and pose space which yields improved reconstructions. Importantly, all random samples from the adapted prior share the same position and orientation with respect to the partial target, greatly simplifying posterior inference.

To address alignment issues, some work has proposed to explicitly calibrate the acquisition devices to standardize pose \citep{fuessinger2018planning,navarro2022reconstruction}. For data created without such standardization, avoiding this problem requires careful joint optimization of pose and shape \citep{madsen2019probabilistic,fuessinger2019virtual,ebert2022reconstruction}. We focus on the latter, making the optimization mostly independent from each other. This issue is sometimes neglected, negatively impacting many applications.

In this paper, we propose a procedure that recalculates the model prior on the fly, modifying the basis to approximate a target-specific model. This leads to improved reconstruction performance as the joint optimization of pose and shape is made simpler. Our method does not require access to the original training data making it applicable in plug-and-play scenarios. We make the following contributions:

\begin{enumerate}
\item We highlight and discuss the importance of a shared rigid alignment between training data and partial target data. 
\item We propose an efficient linear operator that adapts an existing linear model to a target-specific model without requiring access to the original training data. This allows for plug-and-play applications.
\item We experimentally show that the simplified inference problem exhibits significantly improved reconstruction performance using various inference approaches.
\end{enumerate}

Our work is motivated by applications in medical image analysis and forensics. One application is in forensic reconstruction, where the completed shape of a femur bone is predicted to estimate sex and stature \citep{mall2000determination, djorojevic2014morphometric}. Another important application is in osteotomy planning, where surgery is required to correct malunited bones \citep{roth2022accuracy} by inferring the healthy anatomy from the intact portion.

We argue that target-specific prior models should always be used whenever a complete shape is predicted based on partial observations, due to the resulting inference problem being much simpler. For every tested inference method target-specific models exhibit better reconstructive distributions compared to all tested target-agnostic ones. Our method allows to adapt any existing model alignment to the target-specific one making such models always available, even if the training data is not.

\section{Related work}
\label{sec:relatedWork}

In this paper, we focus on point distribution models with an empirically estimated Gaussian distribution for 3D shape reconstruction. PDMs generally separate shape from pose and are one of the simplest models to showcase the alignment problems we are interested in. They are used in various areas \citep{dupraz2022using,shen2012detecting}. Shape models can simplify problems using an explicit latent space \citep{atkins2022prediction,yan2010discrete,bernard2017shape}. Partial targets are also investigated \citep{bernard2017shape,khallaghi2015statistical}. Other PDM based approaches include various localized models \citep{jud2017localized,wilms2017multi,wilms2020kernelized} that move away from the assumption of a single Gaussian distribution. Some methods rely on hierarchical structures \citep{cerrolaza2011multi,zhang2014anatomical}. 

The discretization of the shape is important to PDMs. While we do not discuss this issue, other approaches avoiding this have been explored. One such approach is to use diffeomorphisms to map one shape to another. Early work morphs simple planes \cite{avants2006lagrangian,ashburner2013symmetric} and was later expanded to more complicated domains \cite{lui2012optimization,miller2014diffeomorphometry}. Neural networks are also used to estimate such diffeomorphisms \cite{sun2022topology}.

Apart from PDMs, there are numerous other approaches in statistical shape modelling \citep{heimann2009statistical,bohlender2021survey,el2021high}, i.e. skeleton models \citep{pizer2013nested}, diffeomorphic models \citep{yushkevich2019diffeomorphic}, appearance models \citep{cootes2001statistical}, implicit shape models \citep{yenamandra2021i3dmm} and others \cite{zheng2017statistical}. Other approaches like principal geodesic analysis focus on dimensionality reduction on manifolds \citep{zhang2013probabilistic}. Given a target, learning based shape reconstruction methods approximate the entire shape distribution using a single shape \cite{yu2021pointr,chibane2020implicit}. These mentioned methods can also suffer from a pose misalignment between training and target data. However, there are also models that are invariant to pose and should not be affected by the alignment issues we discuss. Most of them are based on some form of intrinsic measure, like a distance matrix \cite{srivastava2009elastic,reuter2009laplace}.

Besides medical applications, PDMs are used in face modelling for computer vision \citep{li2022comprehensive,egger20203d}. Reconstruction in vision also has a probabilistic formulation \citep{gerig2018morphable}. Differentiable renderers made the wider adaption of neural network based approaches possible \citep{genova2018unsupervised}. Data deficiencies can be reduced with synthetic data created using shape models \citep{wood2021fake}. Shape models are often used to disentangle the parameter space helping generalization \citep{grassal2022neural,zheng2022avatar}. Other applications of shape models include early work on morphometrics \citep{adams2004geometric} or physiological development studies of mice \citep{cates2017shape}.

Next, we focus on the setting of this paper: partial target reconstruction using PDMs. There is work recommending to reconstruct targets by specifically building shape models with the training data properly aligned to the target \cite{albrecht2013posterior,baka2010confidence}. We refer to such models as target-specific models. However, they leave an important gap by not precisely defining the alignment, evaluate the advantages nor describe different strategies to align. 

\paragraph{Target-specific model applications.} 

Training data is appropriately aligned to a specific target. When reconstructing partial shapes \cite{fuessinger2018planning,navarro2022reconstruction}, shared landmarks can be used to create a shape model which approximates a target-specific model. The influence of training data alignment was investigated for partial femur reconstructions \cite{madsen2019probabilistic}. Target-specific models performed much better in finding the correct correspondence between model and partial target.

\paragraph{Target-agnostic model applications.}

In the literature, target-agnostic models are still commonly used. These models are not specifically built for individual partial targets, instead they are often intended for complete targets to, for example, establish correspondence during registration. However, target-agnostic models have been used for cranial reconstructions of synthetic defects using analytical posteriors and elastic iterative closest point methods \cite{fuessinger2019virtual,salhi2020statistical}. They are also used for femur reconstructions \cite{ebert2022reconstruction}. There are also experiments that look at reconstructive quality for different amounts of observed femurs \cite{zhang2017accuracy}. We perform a similar femur reconstruction experiment in this paper. While most reconstruction algorithms have the theoretical capability of providing the correct solution even with target-agnostic models, they often fail short due to a much harder optimization landscape as we will show.

\section{Background}

This Section introduces the shape model building process from unaligned training data. We provide a detailed description of the resulting shape models.

\subsection{Data}

Our training data consists of $n$ registered but unaligned 3D surfaces $\tilde{\Gamma}_1, \ldots, \tilde{\Gamma}_n \subset \mathbb{R}^3$. The shapes are Gaussian distributed with an arbitrary pose in space. The tilde of $\tilde{\Gamma}_i$ indicates random position and orientation.

As the data is registered to some reference surface domain $\Omega \subset \mathbb{R}^3$, we can express all $\tilde{\Gamma}_i$ with an associated deformation field $\tilde{u}_i: \Omega \mapsto \mathbb{R}^3$ mapping each reference point to its corresponding point as $\tilde{\Gamma}_i =  \{x + \tilde{u}_i(x) ~|~ x \in \Omega \}$.

To build a shape model all $\tilde{\Gamma}_i$ are aligned using generalized Procrustes analysis (GPA) \citep{gower1975generalized}. This removes the random rigid transformation of $\tilde{\Gamma}_i$. GPA aligns all shapes $\tilde{\Gamma}_i$ iteratively to their mean shape $\tilde{\Gamma}_M$ defined as 

\begin{equation}
\label{eq:surfaceFromDef}
\tilde{\Gamma}_M = \{x + \frac{1}{n}\sum_i^n \tilde{u}_i(x) ~|~ x \in \Omega \}.
\end{equation}

This is done until convergence. To simplify the notation we update the reference domain $\Omega$ to be equal to $\tilde{\Gamma}_M$. Aligning all shapes on $X \subseteq \Omega$ involves calculating the optimal position and orientation:

\begin{equation}
\begin{split}
\label{eq:gpaAlignmentOperator}
&u_i^X(x) = R_i^X(x + \tilde{u}_i(x) - \mathcal{T}^X[\tilde{u}_i](x)) - x\quad \text {with} \\
&\mathcal{T}^X[\tilde{u}_i](x) = \frac{1}{V_X} \int_X \tilde{u}_i(y) dy \quad \text{and} \quad V_X = \int_X dz.
\end{split}
\end{equation}

The translation operator $\mathcal{T}^X[\tilde{u}_i]$ returns a constant function that maps to the average deformation of $\tilde{u}_i$ on $X$. For that the surface normalization constant $V_X$ is required. 
We consider the resulting fields $u_i^X$ to be aligned on the domain $X$. This definition of the rigid data alignment will be useful in \cref{sec:alignment}. Every such deformation field defines an aligned shape $\Gamma_i^X$ as

\begin{equation}
\label{eq:gammaiOmega}
    \Gamma_i^X = \{x + u_i^X(x) ~|~ x \in \Omega \}.
\end{equation}

The definition of $\Gamma_i^X$ is a generalization of the special case $X=\Omega$ which leads to $\Gamma_i^\Omega$. In the literature most shape models are built using $\Gamma_i^\Omega$.

\subsection{Shape models}

We refer to shape models built from $\Gamma^X_i$ as aligned on $X$. All definitions also apply to the special case of $X=\Omega$.

We use Gaussian process morphable models (GPMM) \citep{luthi2017gaussian} as the mathematical formulation for describing shape variation. GPMM is a continuous generalization of the classical PDM. Following \cref{eq:gammaiOmega}, shapes are uniquely defined by their deformation field. We define a distribution over deformation fields using an empirically estimated Gaussian process $\GP$:

\begin{equation}
    u^X \sim \GP(\mu^X, k^X),
\end{equation}

where $u^X$ is a sample deformation field from the GPMM. The mean $\mu^X$ is the sample mean and $k^X$ is the sample covariance.

As the model is built from $n$ training shapes it has at most a rank of $r\leq n-1$. Similar to principal component analysis, there is a continuous low dimensional representation of $k^X$. A convenient basis of $k^X$ is given by the Karhunen-Loève expansion which can be calculated using for example the Nyström \citep{drineas2005nystrom} or pivoted Cholesky \citep{dolz2019error} method. The covariance function $k^X$ can be written as a linear combination of its basis functions:

\begin{equation}
\label{eq:klSum}
u^X[\boldsymbol\alpha](x) \sim \mu^X(x) + \sum_{i=1}^r \alpha_i \sqrt{\lambda_i^X}{\phi_i^X(x)},
\end{equation}

where $u^X:\mathbb{R}^r\mapsto \Omega \mapsto\mathbb{R}^3$ and $\boldsymbol\alpha = (\alpha_i)_i$ with $\alpha_i \in N(0, 1)$. The pairs $(\lambda_i, \phi_i)$ are the eigenvalues and eigenvectors of the corresponding covariance operator 

\begin{equation}
K^X[f](\cdot) = \int_{\Omega}k^X(\cdot, x)f(x)\, dx,
\end{equation}

where $K^X:(\Omega\mapsto\mathbb{R}^3)\mapsto\Omega\mapsto\mathbb{R}^3$.

The low rank representation reduces computational complexity and limits the influence of observed noise, akin to probabilistic principal component analysis \citep{tipping1999probabilistic}.

Such empirical shape models $\GP(\mu^X, k^X)$ are built under the specific data alignment objective in \cref{eq:gpaAlignmentOperator}. Even with infinite data there is still a nullspace spanned by the derivatives of the position and orientation alignment. This can be seen by reformulating \cref{eq:gpaAlignmentOperator} as a minimization objective. This means that every deformation field $u^X$ has an average deformation of zero on $X$, meaning the position of the part $X$ remains stationary.

\subsubsection{Issues during partial target reconstruction}
\label{sec:issuesReco}

A common approach to reconstruct partial targets $\tilde{\Gamma}_\tau$ observed on $X\subseteq \Omega$ is to align them to the model mean on $X$ while the model $\GP(\mu^\Omega, k^\Omega)$, aligned on $\Omega$, is used to reconstruct the target. 
This is problematic. Following \cref{eq:gpaAlignmentOperator} and the definition of empirical shape models, there is a clear mismatch in underlying distributions. This can happen to various degrees depending on how much the observed and the alignment domains differ. 
We refer to the model as target-specific only if the target is aligned to the model mean on the same domain that the training data was aligned to. Otherwise, we call it target-agnostic.

\begin{figure}
    \centering
    \includegraphics[width=0.6\linewidth]{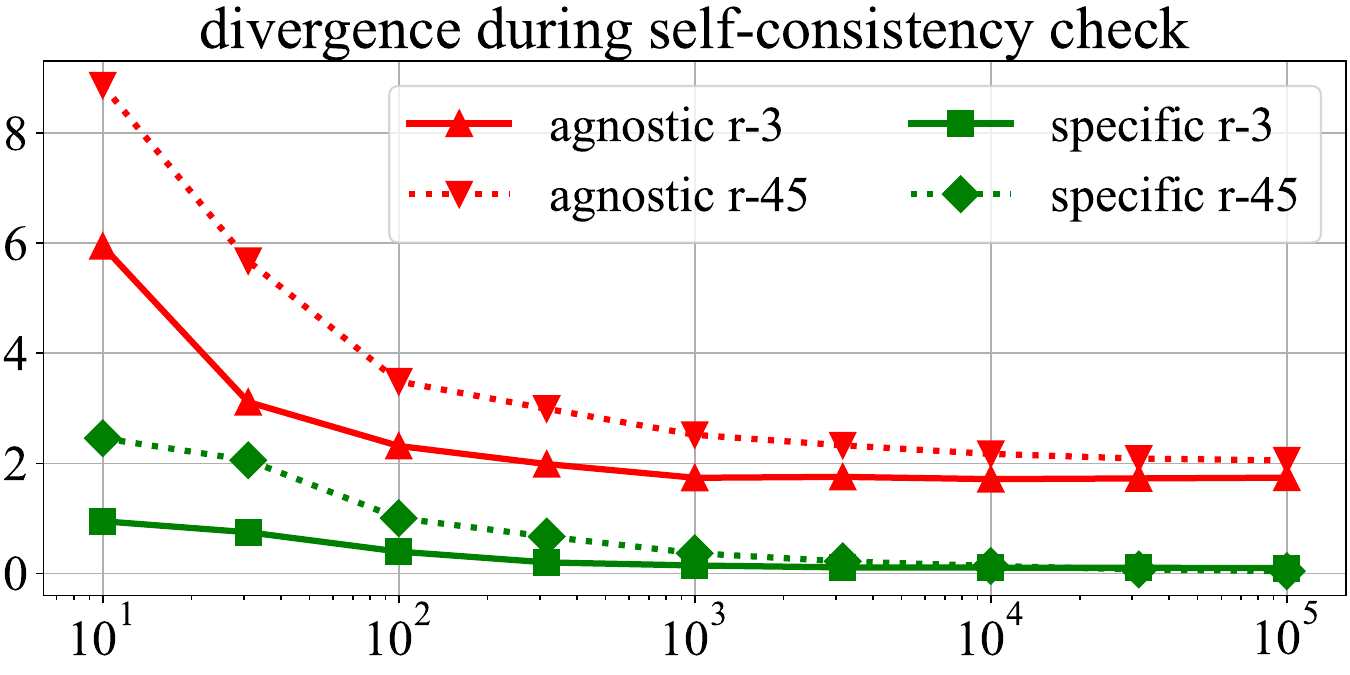}
    \caption{The $x$-axis shows the number of random targets and the $y$-axis shows the symmetric KL divergence to the prior for rank 3 and 45. Target-agnostic models cannot recover the prior accurately due to biased reconstructions.}
    \label{fig:divergence}
\end{figure}

To showcase the consequences of target-agnostic models we perform a self consistency check \cite{talts2018validating}. The idea is to sample from the prior shape models and mask these samples to form partial targets. Each such target has a posterior distribution of valid reconstruction. In theory, aggregating correctly estimated posterior distributions would converge to the prior distribution. This gives rise to a simple test of the models used for reconstructions. If the prior cannot be recovered, then the estimated posterior distributions are biased. \Cref{fig:divergence} shows how target-agnostic models do not successfully recover the prior, even in lower dimensions.

\subsection{Full posterior formulation}

In practice, the correspondence between the model and target is often unknown, making pure shape regression with posterior shape models insufficient. This paper focuses on the setting of defining a parameter distribution of possible reconstructions of a given partial target $\Gamma_\tau$ without having the correct correspondence. 

Formally, we are interested in the distribution of shape parameters $\boldsymbol\alpha$, rigid transformation parameters $\boldsymbol r = [r_\psi ~ r_\gamma ~ r_\zeta]$ and $\boldsymbol t = [t_x ~ t_y ~ t_z]$ that explain $\Gamma_\tau$ well. As such, the formulated posterior depends on what shape model prior is used, which we indicate by denoting the alignment domain. With the parameters $\theta = [\boldsymbol\alpha, \boldsymbol r, \boldsymbol t]^T$ the posterior of $\theta$ is

\begin{equation}
\label{eq:fullFormulation}
p^X(\theta | \Gamma_\tau) \propto p^X(\Gamma_\tau|\theta)p(\theta).
\end{equation}

Each $\theta$ corresponds to a complete shape $\Gamma[\theta](x) = R_{\boldsymbol r}(x+u^X[\boldsymbol\alpha](x))+\boldsymbol t$. The prior $p(\theta)$ is independent for each component. For shape it is $\boldsymbol\alpha \sim \mathcal{N}(0,I)$. We assume a uniform prior for the rotation and translation parameters. 
For rotation with Eulerian angles, this equates to $U(0,2\pi)$. For translation, large bounds $U(-B,B)$ are used to avoid an improper prior. We omit the superscript for the posterior $p(\theta | \Gamma_\tau)$ if the used shape model is unimportant.

As the likelihood $p^X(\Gamma_\tau|\theta)$, we assume the difference between corresponding points on the model shape $\Gamma[\theta]$ and $\Gamma_\tau$ to be Gaussian and independently distributed. For this, we assume some function $CP(x, \Gamma_\tau)$ that returns the corresponding point on $\Gamma_\tau$ of $x \in \Omega$. See \cref{sec:appendixLikelihood} for a more detailed description.

As the true correspondence function $CP$ is often unknown, estimates are employed. We will later use the common closest point approximation. For any point $x$ the closest point on $\Gamma_\tau$ is defined as 

\begin{equation}
\label{eq:closestPoint}
    CLP(x, \Gamma_\tau) = \argmin_{x' \in \Gamma_\tau} \norm{x - x'}.
\end{equation}

The main idea behind $CLP$ is that if two surfaces are close to each other, then $CP$ is well approximated.

How the data was rigidly aligned in \cref{eq:gpaAlignmentOperator} and subsequently what shape model is used critically influences the posterior $p^X(\theta | \Gamma_\tau)$ and how complicated it is to sample from the distribution. We will showcase this influence using some standard inference methods in \cref{sec:fullInference}. 

Similar to the shape models, we also refer to these resulting posteriors as target-specific or target-agnostic with the shortened notation $p^X$ and $p^\Omega$, respectively.

\section{Method}
\label{sec:alignment}

To introduce the method in this Section we use a simplified GPA alignment, restricted to aligning the position. That is we assume $\tilde{u}_i$ are shapes with no additional rotations as noise. In \cref{sec:alignmentRot} and after we use the full GPA alignment.

Our goal is to define a simple operator that maps an input model with arbitrary alignment to a desired alignment on $X \subseteq \Omega$. This would give a lot of flexibility when reconstructing varying partial targets and using prebuilt models. We will begin defining such an operator for a single shape. We exploit that the positional alignment can be directly formulated as a projection operator. Using the identity operator $\mathcal{I}$ and omitting the rotation in \cref{eq:gpaAlignmentOperator} leads to

\begin{equation}
\begin{split}
u_i^X(x) &= \tilde{u}_i(x) - \mathcal{T}^X[\tilde{u}_i](x) \\
&= (\mathcal{I}[\tilde{u}_i] - \mathcal{T}^X[\tilde{u}_i]) (x) \\
&= \mathcal{P}^X[\tilde{u}_i](x).
\end{split}
\end{equation}

The operator $\mathcal{P}^X$ is a projection and is the complement of $\mathcal{T}^X$. It removes the average deformation on $X$ from a given deformation field $\tilde{u}_i$. This projection $\mathcal{P}^X$ has as image the shape space of $\GP(\mu^X, k^X)$ and as kernel the space of constant deformation fields. That means it only ever changes the position of the modified shapes and can be understood as moving a fixed shape between differently aligned shape spaces by changing the position.

A big advantage of linear operators is that we can apply $\mathcal{P}^X$ to a sample mean and covariance function instead of individual shapes \cite{sarkka2011linear}. This allows us to project an existing arbitrary model $\GP(\mu^Z, k^Z)$ aligned on $Z \subseteq \Omega$ independent of what $Z$ is. We define a new model using the projected sample mean $\mathcal{P}^X[\mu^Z](x)$ and projected sample covariance function
\begin{equation}
\label{eq:projectedKernel}
\sum_{i=1}^r \alpha_i ~ \lambda_i^Z ~ \mathcal{P}^X[\phi_i^Z](x) ~ \mathcal{P}^X [\phi_i^Z](x')^T.
\end{equation}

We will write $\GP(\mathcal{P}[\mu^Z], \mathcal{P}^X k^Z \mathcal{P}^{X^T})$, where $\mathcal{P}^X k^Z \mathcal{P}^{X^T}$ represents the left and right applied operator as seen in \cref{eq:projectedKernel}. More compactly, we further denote the associated posterior as $p^\mathcal{P}$. 
All the projected deformation fields will have the same positional alignment as if they were part of $\GP(\mu^X, k^X)$. For the simplified alignment setting, the following equality holds:

\begin{equation}
    \label{eq:keqpkp}
    \mu^X = \mathcal{P}^X[\mu^Z], ~~~~ k^X = \mathcal{P}^X k^Z \mathcal{P}^{X^T}.
\end{equation}

This formalism of applying a projection operator to an existing model fulfills all our requirements as it is agnostic to $Z$ and leads to the model aligned on $X$.

\begin{algorithm}
\caption{Projection to create target-specific model}\label{alg:proj}
\begin{algorithmic}[1]
\Statex Input: low rank $\GP$: $\mu^Z$, $(\lambda_i^Z)_i$, $(\phi_i^Z)_i ~~\text{and}~ X$
\State $\boldsymbol\mu^Z, (\boldsymbol\phi_i^Z)_i \gets  \Call{discretize}{\mu^Z, (\phi_i^Z)_i}$ \Comment{discr. on $\Omega$}
\State $M \gets [\boldsymbol\phi_{t_x} ~ \boldsymbol\phi_{t_y} ~ \boldsymbol\phi_{t_z}]$ \Comment{$M \in \mathbb{R}^{3N \times 3}$}
\State $M_X \gets M(1\ldots3N_X,::)$ \Comment{$X$ as starting indices}
\Function{projectVector}{$\boldsymbol\phi$} 
\Statex \Comment{avoid explicit projection matrix $P^X$}
\State \Return $\boldsymbol \phi - M(M_X^T M_X)^{-1}M_X^T\boldsymbol\phi(1\ldots3N_X)$
\EndFunction
\State $(\boldsymbol\phi_i^X)_i \gets$ \Call{projectVector}{$(\boldsymbol\phi_i^Z)_i$}
\State $\boldsymbol\mu^X \gets$ \Call{projectVector}{$\boldsymbol\mu^Z$} \Comment{optional}
\State $(\lambda_i^X)_i, (\boldsymbol\phi_i^X)_i \gets \Call{diagonalizeBasis}{(\lambda_i^Z)_i, (\boldsymbol\phi_i^X)_i}$ 
\Statex \Comment{i.e. SVD on Gram matrix of $(\boldsymbol\phi_i^X)_i$}
\State $\mu^X, (\phi_i^X)_i \gets \Call{interpolate}{\boldsymbol\mu^X, (\boldsymbol\phi_i^X)_i}$ 
\State \Return low rank $\GP$: $\mu^X$, $(\lambda_i^X)_i$, $(\phi_i^X)_i$
\end{algorithmic}
\end{algorithm}

In \Cref{alg:proj} we list a discrete implementation of $\mathcal{P}^X$. The domain $\Omega$ is discretized to $N$ points in total, and $N_X$ points on the observed domain $X$. The matrix $M$ is a basis of the projection kernel and is built from normalized discretizations of the translations. It contains for example the normalized vector for the $x$-translation
\begin{equation}
\label{eq:translationEigf}
\boldsymbol\phi_{t_x} = \frac{1}{\sqrt{N}} [1,0,0,1,0,\ldots]^T.
\end{equation}

This realignment has a computational complexity of $O(Nr^2)$, where $r$ is the rank of the input model. The highest complexity stems from a matrix multiplication during the orthogonalization of the intermediate basis $\{ \boldsymbol\phi_i \}_i$.

\subsection{Approximating non-linear effects like rotations}
\label{sec:alignmentRot}

Data is not only normalized in linear ways, i.e. the non-linear orientation is also normalized. This poses a challenge for realigning an existing model as the model space itself is linear. However, some non-linear aspects allow for a good linear approximation.

The idea is to adapt $\mathcal{P}^X$ by extending the kernel of the projection for non-linear data normalization. We linearly approximate some scalar-parametrized $A[\vartheta]$ around the mean, here at $\vartheta=0$, by extending the projection kernel with the derivative of $A[\vartheta]$ at $\vartheta=0$:

\begin{equation}
    a(x) = \pdv{A[\vartheta]}{\vartheta}\Bigr|_{\substack{\vartheta=0}}(x).
\end{equation}

This is repeated for each degree of freedom of the data normalization. For shape models there is often already an explicit parametrization of the data normalization making this approach well suited.

In our application, the orientation allows for a straightforward parametrization with three orthogonal rotation axis $R_\psi[\vartheta], R_\gamma[\vartheta], R_\zeta[\vartheta]$ through the center of $X$. The derivatives of these rotation operators are used to extend the kernel of $\mathcal{P}^X$, which is straightforward for \cref{alg:proj}. These derivatives are discretized and normalized to extend $M$ to $[\boldsymbol\phi_{t_x} ~ \boldsymbol\phi_{t_y} ~ \boldsymbol\phi_{t_z} ~ \boldsymbol\phi_{r_\psi} ~ \boldsymbol\phi_{r_\gamma} ~ \boldsymbol\phi_{r_\zeta}]$.

The error of the linear approximation is entirely dependent on $A[\vartheta]$. It can be arbitrarily bad. However, it is known that a linear approximation for small rotations works well as the error scales with the distance to axis and the sinus of the rotation angle. 
For a further investigation of the error scaling and a visualization of the effects of the projection operator see \cref{sec:hingeAppendix}. 
In our applications the approximation lead to negligible errors. 

In practice, we do not project the mean as this can add additional errors with little benefit. In the notation, we continue to include the projected mean for conceptual clarity.

This extended alignment leads to shape models that are approximately invariant to rotation perturbations and invariant to translation observed on $X$. Of course, defining $a(\cdot)$ works for any linear $A[\vartheta]$, and it can easily be checked that position normalization leads to $[\boldsymbol\phi_{t_x} ~ \boldsymbol\phi_{t_y} ~ \boldsymbol\phi_{t_z}]$ for \cref{alg:proj}. Other linear extensions like scaling or shearing are similarly simple.

\section{Experiments}
\label{sec:fullInference}

To verify the importance of the model priors, we examine how the alignment influences the inference quality for target-agnostic and target-specific alignments when performing reconstructions of partial targets with different inference approaches. The inference methods vary in computational cost to cover a broad range of settings with different constraints.

The three inference approaches we investigate are a fast and cheap EM-like optimization, sampling using a medium-sized computational budget, and a computationally more expensive variational approximation method using a GPU. All of these methods have numerous variants in the literature. The code for the experiments is publicly available\footnote{\url{https://github.com/unibas-gravis/posterior-shape-models-revisited}}.

\paragraph{Non-rigid iterative closest point} A simple method to obtain a single reconstruction is the iterative closest point (ICP) algorithm \citep{besl1992method,zhang1994iterative}. It is still used due to its simplicity and how quickly it converges. In the original paper on ICP, a rigid transformation was computed based on $CLP$. However, non-rigid ICP (nICP) is also possible \citep{allen2003space,amberg2007optimal}. We formulate nICP as finding the maximum of $p(\theta | \Gamma_\tau)$ defined in \cref{eq:fullFormulation} by iteratively optimizing shape and then pose. The name originates from the use of $CLP$ from \cref{eq:closestPoint} to fix correspondence making the individual optimizations possible. However, the single reconstruction of nICP does not quantify uncertainty and as an EM-like method it can converge to local minima.

\paragraph{Metropolis Hastings}

Approximate inference can be performed by using a sampling-based approach. The Metropolis Hastings (MH) algorithm \citep{robert2010metropolis} has theoretical guarantees as a Markov chain Monte Carlo approach. We employ a mixture of block-wise random walk proposals \citep{schonborn2017markov}. To improve mixture characteristics we add a symmetric version of a proposal based on Gaussian process regression \citep{madsen2020closest} to the isotropic shape proposals. The sample $\theta$ with the highest $p(\theta | \Gamma_\tau)$ is used as MAP estimate to calculate the reconstruction error, while the predicted variance is calculated as the empirical variance of the shapes $\Gamma[\theta]$ of the Markov chain.

\paragraph{Normalizing flows} A popular group of variational inference methods are normalizing flows \cite{rezende2015variational,papamakarios2021normalizing}. Bijections are used to transform a simple shape parameter distribution $q(\theta)$ into $q^*(\theta) \approx p(\theta | \Gamma_\tau)$, where the KL-divergence to $p(\theta | \Gamma_\tau)$ is the loss. We use eight layers of iResBlocks \cite{perugachi2021invertible}. The empirical variance is calculated from a number of samples $\Gamma[\theta]$ mapped through the flow, while the reconstruction error is determined using the annealed system \cite{van1987simulated} approximating the MAP of $p(\theta | \Gamma_\tau)$.

\subsection{Femur reconstruction}
\label{sec:syntheticRecon}

We perform a leave-one-out experiment for femurs and reconstruct them by using the distribution $p(\theta | \Gamma_\tau)$ from \cref{eq:fullFormulation}. We compare $p^\Omega$, $p^X$ and our $p^\mathcal{P}$. For our model we include the rotation approximation from \cref{sec:alignmentRot}. We optimize over shape and pose using $CLP$ as approximation of the true correspondence. The target-specific model uses the ground truth $X$ for alignment, while our method uses an estimate for $X$. More details on the setup are in \cref{sec:detailedResults}

We use a femur dataset\footnote{https://shapemodelling.cs.unibas.ch/femur-reconstruction-challenge/register} consisting of 47 full femurs that was first published on the SICAS medical image repository. The left out femurs are cut to represent a partial target. In \cref{sec:scaphoids}, we perform a similar test with a scaphoid dataset.

\begin{figure}
    \centering
\includegraphics[width=0.4\linewidth]{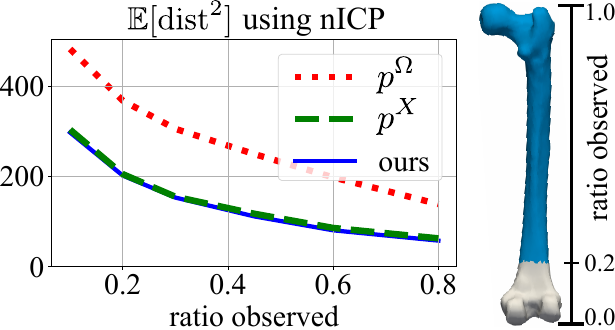}
    \caption{The right illustrates in white how much of the femur is observed for the reconstructions seen on the left, where the $y$-axis lists the squared distance to the ground truth femur. Target-agnostic models have significantly higher expected distance compared to the similar performing target-specific and our models.}
    \label{fig:icpPost}
\end{figure}

\begin{figure*}[!t]%
    \centering
    \includegraphics[width=\linewidth]{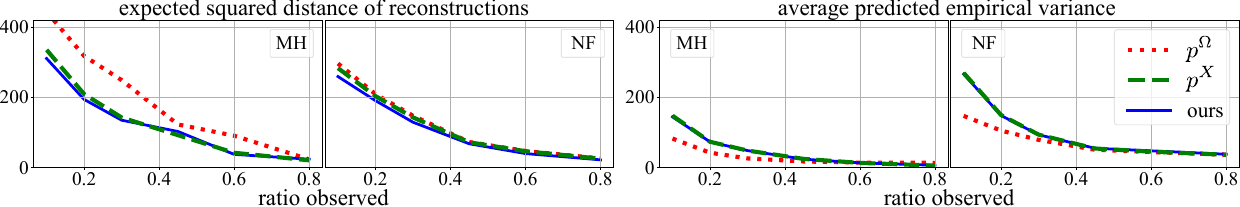}%
    \caption{Results of the femur reconstruction experiment using MH and normalizing flows. 
    Target-specific $p^X$ and our projected distributions $p^\mathcal{P}$ lead to significantly better reconstructions and more adequate predicted variance compared to target-agnostic distributions $p^\Omega$.}%
    \label{fig:mhPost}%
\end{figure*}

\begin{table*}
\begin{center}
\setlength\tabcolsep{4pt}
\resizebox{\textwidth}{!}{%
\begin{tabular}{|l||c|c|c|c|c||c|c|c|c|c||c|c|c|c|c||}
\hline
method & \multicolumn{5}{c||}{non-rigid ICP} & \multicolumn{5}{c||}{Metropolis Hastings} & \multicolumn{5}{c||}{normalizing flow}  \\
\hline
ratio observed & 20\% & 30\% & 45\% & 60\% & 80\% & 20\% & 30\% & 45\% & 60\% & 80\% & 20\% & 30\% & 45\% & 60\% & 80\% \\ 
\hline
$p^\Omega(\theta | \Gamma_\tau)$  & 368&305&249&198&139 & 318&249&122&90&23 & 209&147&72&47&25 \\
$p^X(\theta | \Gamma_\tau)$  & 205&155&117&86&\tb{63} & 209&\tb{142}&\tb{92}&42&\tb{20} &203&143&72&46&25 \\ \hline
ours $p^\mathcal{P}(\theta | \Gamma_\tau)$ & \tb{202}&\tb{153}&\tb{115}&\tb{85}&\tb{63} &
\tb{201}&143&103&\tb{37}&23 &\tb{190}&\tb{129}&\tb{67}&\tb{39}&\tb{21}\\
\hline
\end{tabular}}
\end{center}
\caption{Mean squared error to the ground truth target using the correct correspondence when using nICP, MH, and normalizing flow inference methods. The metric is listed averaged over the full domain $\Omega$. Lowest error in bold for each column. Target-agnostic \cite{fuessinger2019virtual,salhi2020statistical,ebert2022reconstruction,zhang2017accuracy} and target-specific \cite{baka2010confidence,albrecht2013posterior,madsen2019probabilistic} models are each widely used even though the target-specific $p^X$ provides better reconstructions.}
\label{tab:expTableShort}
\end{table*}

For the nICP algorithm, we use $150$ iterations. Since it is a computationally inexpensive method, this takes at most one minute on a single CPU core. For MH, we use $50$k samples with $5$k burn-in, requiring approximately $30$ minutes on a single CPU core. Normalizing flows involve a more expensive setup, taking around $1.5$ hours on an A100 GPU per reconstruction, amounting to over $400$ hours of GPU computation time per tested posterior distribution $p(\theta | \Gamma_\tau)$.

\Cref{fig:icpPost} shows the single-reconstruction results using nICP, while \cref{fig:mhPost} presents results for MH and normalizing flows. We report the average squared distance based on the true correspondence function $CP$, along with the empirical variance of the generated samples where applicable.

The reconstruction error for all methods is higher when using a target-agnostic $p^\Omega$ compared to the target-specific $p^X$ or our projected $p^\mathcal{P}$, both of which achieve similar performance. With the increased computational budget of normalizing flows, the reconstructive quality using $p^\Omega$ nearly reaches the level of $p^X$. Additionally, using $p^\Omega$ leads to lower predicted variance across all tested methods.

In summary, using the target-agnostic $p^\Omega$ leads to unnecessarily biased reconstructions and underestimated variance. We recommend using the target-specific $p^X$ when available or $p^\mathcal{P}$ via our projection when training data is unavailable. This holds even if a significant amount, here 80\%, of the full shape is observed.

\subsection{Skull osteotomy}
\label{sec:application}

\begin{figure}
    \centering
\includegraphics[width=0.7\linewidth]{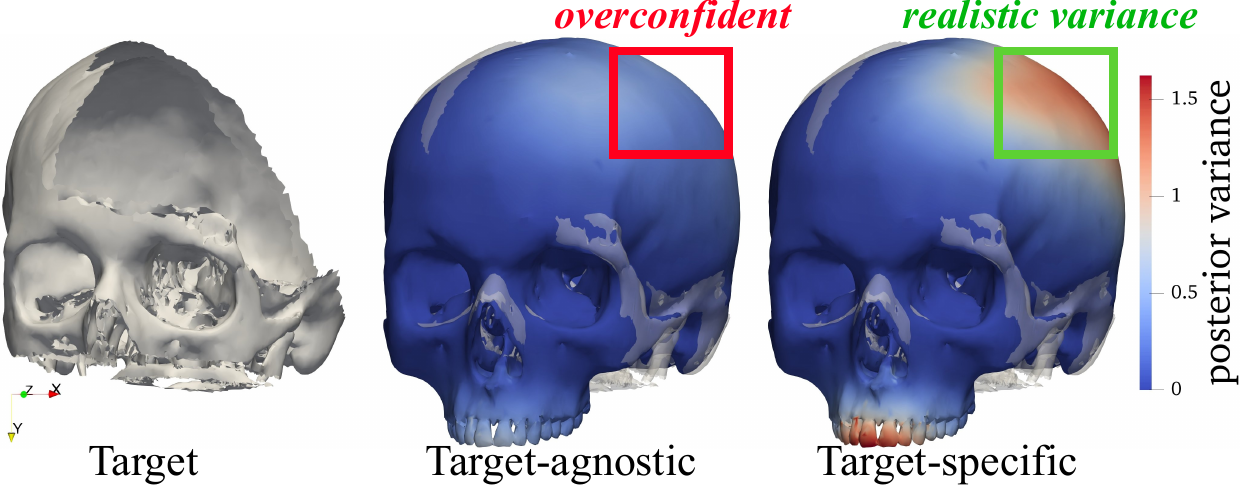}
    \caption{The predictive posterior is imposed over the target. The target-specific model leads to a more appropriate uncertainty prediction for the unobserved parts.}
    \label{fig:skullVariance}
\end{figure}

A common application in the medical setting is a healthy reconstruction of a pathology. This relies on an outlier model or removing the affected part and then reconstructing it. We showcase surgery planning for an osteotomy where large parts of the skull are removed as they are affected by a pathology. We use 46 healthy skull shapes to build the shape model originating from an internal dataset. To demonstrate that our method can not only be used for purely empirical shape models, we further localize the model following \cite{wilms2020kernelized}. For more model details see \cref{sec:detailedModelSkull}.

These kinds of experiments lack a ground truth for comparison. While \Cref{sec:syntheticRecon} provides quantitative results, this skull experiment shows a qualitative example.

We provide a single reconstruction using the nICP algorithm. This shape can be used to fix correspondence which gives access to the predictive posterior for the shape parameter using posterior shape models. This is calculated using the target-agnostic and our projected model. The resulting posteriors are shown in \cref{fig:skullVariance}. For a fair comparison, the same correspondence is used for both models.

We showed qualitatively that our model has a higher uncertainty in the unobserved regions compared to the target-agnostic model. This mirrors our findings in \cref{sec:fullInference} for a more complex shape with more observation noise.

 


\section{Discussion}

Our experiments focus on empirical point distribution models. Our projection method is applicable to models with a linear basis, including local linear bases, though this loses strict localization. Our method also applies to models defined or enhanced by an analytical kernel.

Our approach excludes pose relative to the observed domain from the shape space. However, one can also include the pose variance in the model space \cite{albrecht2013posterior}. Such an approach would converge to ours in the infinite limit of added pose variance, as it introduces an additional distribution over pose in the shape space. This looses the clear distinction between shape and pose, while our approach explicitly separates shape and pose without requiring an infinite limit.

\subsection{Best practices}
With this work, we aim to establish a new set of best practices for partial data reconstruction. The intended setting of our recommendations consists of a shape space with mainly low-frequency components resulting in smooth shapes, with either no prior or a uniform prior over pose. The recommendations apply to both single reconstructions and to approximate inference. We formulate our recommendations with no specific framework in mind.

\begin{enumerate}
    \item Gather training data or obtain a model $\GP(\mu^\Omega, k^\Omega)$.
    \item If $X$ is unknown, then approximate it. Initial estimates can be obtained using a rough rigid alignment and a simple heuristic.
    \item Formulate the target-specific $p^X$ or project $\GP(\mu^\Omega, k^\Omega)$ using $\mathcal{P}^X$ to get $p^\mathcal{P}$ as outlined in \cref{sec:alignment}.
    \item If correspondence is known, use posterior shape models. Otherwise, use an inference method on $p^X$. For iterative methods like nICP or MH, updates of $X$ every few iterations are simple by going back to step 2.
\end{enumerate}

\section{Conclusions and limitations}

Various data alignment strategies are used in the literature. We identify two groups: target-agnostic and target-specific alignment. In a Bayesian reconstruction approach, these alignment strategies modify the training data defining the prior distribution, which in turn influences the posterior of reconstructions. The choice of alignment impacts the feasibility of variational approximations and sampling techniques for the posteriors. This paper examines these effects and establishes best practices for alignment.

In a leave-one-out femur reconstruction experiment, we show that target-specific alignment consistently improves reconstruction accuracy and variance estimation across all tested inference methods. In contrast, the commonly used target-agnostic alignment leads to suboptimal reconstructions. Additional experiments on two other datasets demonstrate that these findings extend beyond long bones.

Additionally, we formalize the various data alignments as operators normalizing the pose of the data. The image of an operator corresponds to a shape space. Using a linear approximation of the operator equates to a projection and allows to directly realign a model basis to any desired data alignment. However, normalizing non-linear aspects like orientation by projection requires a sufficiently accurate linear approximation, otherwise approximation errors become significant. Since the projection can only be directly built into a linear basis, it is less suitable for non-linear shape models. Nonetheless, proper data alignment remains crucial even for non-linear models.

This projection is particularly useful for end users who do not have access to the original training data due to data privacy regulations. By modifying any pre-built model basis to match a desired alignment, the projection enables modifying existing pipelines. This plug-and-play capability greatly benefits many shape model applications.

\bibliographystyle{unsrtnat}
\bibliography{references}  

\begin{thebibliography}{69}
\providecommand{\natexlab}[1]{#1}
\providecommand{\url}[1]{\texttt{#1}}
\expandafter\ifx\csname urlstyle\endcsname\relax
  \providecommand{\doi}[1]{doi: #1}\else
  \providecommand{\doi}{doi: \begingroup \urlstyle{rm}\Url}\fi

\bibitem[Albrecht et~al.(2013)Albrecht, L{\"u}thi, Gerig, and
  Vetter]{albrecht2013posterior}
Thomas Albrecht, Marcel L{\"u}thi, Thomas Gerig, and Thomas Vetter.
\newblock Posterior shape models.
\newblock \emph{Medical image analysis}, 17\penalty0 (8):\penalty0 959--973,
  2013.

\bibitem[Fuessinger et~al.(2018)Fuessinger, Schwarz, Cornelius, Metzger, Ellis,
  Probst, Semper-Hogg, Gass, and Schlager]{fuessinger2018planning}
Marc~Anton Fuessinger, Steffen Schwarz, Carl-Peter Cornelius, Marc~Christian
  Metzger, Edward Ellis, Florian Probst, Wiebke Semper-Hogg, Mathieu Gass, and
  Stefan Schlager.
\newblock Planning of skull reconstruction based on a statistical shape model
  combined with geometric morphometrics.
\newblock \emph{International journal of computer assisted radiology and
  surgery}, 13\penalty0 (4):\penalty0 519--529, 2018.

\bibitem[Navarro-Jim{\'e}nez et~al.(2022)Navarro-Jim{\'e}nez, Aguado, Bazin,
  Albero, and Borzacchiello]{navarro2022reconstruction}
Jos{\'e}~M Navarro-Jim{\'e}nez, Jos{\'e}~V Aguado, Gr{\'e}goire Bazin, Vicente
  Albero, and Domenico Borzacchiello.
\newblock Reconstruction of 3d surfaces from incomplete digitisations using
  statistical shape models for manufacturing processes.
\newblock \emph{Journal of Intelligent Manufacturing}, pages 1--14, 2022.

\bibitem[Madsen et~al.(2019)Madsen, Vetter, and
  L{\"u}thi]{madsen2019probabilistic}
Dennis Madsen, Thomas Vetter, and Marcel L{\"u}thi.
\newblock Probabilistic surface reconstruction with unknown correspondence.
\newblock In \emph{Uncertainty for Safe Utilization of Machine Learning in
  Medical Imaging and Clinical Image-Based Procedures}, pages 3--11. Springer,
  2019.

\bibitem[Fuessinger et~al.(2019)Fuessinger, Schwarz, Neubauer, Cornelius, Gass,
  Poxleitner, Zimmerer, Metzger, and Schlager]{fuessinger2019virtual}
Marc~Anton Fuessinger, Steffen Schwarz, Joerg Neubauer, Carl-Peter Cornelius,
  Mathieu Gass, Philipp Poxleitner, Ruediger Zimmerer, Marc~Christian Metzger,
  and Stefan Schlager.
\newblock Virtual reconstruction of bilateral midfacial defects by using
  statistical shape modeling.
\newblock \emph{Journal of Cranio-Maxillofacial Surgery}, 47\penalty0
  (7):\penalty0 1054--1059, 2019.

\bibitem[Ebert et~al.(2022)Ebert, Rahbani, L{\"u}thi, Thali, Christensen, and
  Fliss]{ebert2022reconstruction}
Lars~C Ebert, Dana Rahbani, Marcel L{\"u}thi, Michael~J Thali, Angi~M
  Christensen, and Barbara Fliss.
\newblock Reconstruction of full femora from partial bone fragments for
  anthropological analyses using statistical shape modeling.
\newblock \emph{Forensic Science International}, 332:\penalty0 111196, 2022.

\bibitem[Mall et~al.(2000)Mall, Graw, Gehring, and
  Hubig]{mall2000determination}
Gita Mall, Matthias Graw, Kristina-D Gehring, and Michael Hubig.
\newblock Determination of sex from femora.
\newblock \emph{Forensic science international}, 113\penalty0 (1-3):\penalty0
  315--321, 2000.

\bibitem[Djorojevic et~al.(2014)Djorojevic, Rold{\'a}n, Garc{\'\i}a-Parra,
  Alem{\'a}n, and Botella]{djorojevic2014morphometric}
Mirjana Djorojevic, Concepci{\'o}n Rold{\'a}n, Patricia Garc{\'\i}a-Parra,
  Inmaculada Alem{\'a}n, and Miguel Botella.
\newblock Morphometric sex estimation from 3d computed tomography os coxae
  model and its validation in skeletal remains.
\newblock \emph{International journal of legal medicine}, 128:\penalty0
  879--888, 2014.

\bibitem[Roth et~al.(2022)Roth, van Es, Kraan, Eygendaal, Colaris, and
  Stockmans]{roth2022accuracy}
Kasper Roth, Eline van Es, Gerald Kraan, Denise Eygendaal, Joost Colaris, and
  Filip Stockmans.
\newblock Accuracy of 3d corrective osteotomy for pediatric malunited both-bone
  forearm fractures.
\newblock \emph{Children}, 10\penalty0 (1):\penalty0 21, 2022.

\bibitem[Dupraz et~al.(2022)Dupraz, Bollinger, Deckx, Schierjott, Utz, and
  Jacobs]{dupraz2022using}
Ingrid Dupraz, Arthur Bollinger, Julien Deckx, Ronja~Alissa Schierjott, Michael
  Utz, and Marnic Jacobs.
\newblock Using statistical shape models to optimize tka implant design.
\newblock \emph{Applied Sciences}, 12\penalty0 (3):\penalty0 1020, 2022.

\bibitem[Shen et~al.(2012)Shen, Fripp, M{\'e}riaudeau, Ch{\'e}telat, Salvado,
  Bourgeat, Initiative, et~al.]{shen2012detecting}
Kai-kai Shen, Jurgen Fripp, Fabrice M{\'e}riaudeau, Ga{\"e}l Ch{\'e}telat,
  Olivier Salvado, Pierrick Bourgeat, Alzheimer's Disease~Neuroimaging
  Initiative, et~al.
\newblock Detecting global and local hippocampal shape changes in alzheimer's
  disease using statistical shape models.
\newblock \emph{Neuroimage}, 59\penalty0 (3):\penalty0 2155--2166, 2012.

\bibitem[Atkins et~al.(2022)Atkins, Agrawal, Mozingo, Uemura, Tokunaga, Peters,
  Elhabian, Whitaker, and Anderson]{atkins2022prediction}
Penny~R Atkins, Praful Agrawal, Joseph~D Mozingo, Keisuke Uemura, Kunihiko
  Tokunaga, Christopher~L Peters, Shireen~Y Elhabian, Ross~T Whitaker, and
  Andrew~E Anderson.
\newblock Prediction of femoral head coverage from articulated statistical
  shape models of patients with developmental dysplasia of the hip.
\newblock \emph{Journal of Orthopaedic Research{\textregistered}}, 40\penalty0
  (9):\penalty0 2113--2126, 2022.

\bibitem[Yan et~al.(2010)Yan, Xu, Turkbey, and Kruecker]{yan2010discrete}
Pingkun Yan, Sheng Xu, Baris Turkbey, and Jochen Kruecker.
\newblock Discrete deformable model guided by partial active shape model for
  trus image segmentation.
\newblock \emph{IEEE transactions on Biomedical Engineering}, 57\penalty0
  (5):\penalty0 1158--1166, 2010.

\bibitem[Bernard et~al.(2017)Bernard, Salamanca, Thunberg, Tack, Jentsch,
  Lamecker, Zachow, Hertel, Goncalves, and Gemmar]{bernard2017shape}
Florian Bernard, Luis Salamanca, Johan Thunberg, Alexander Tack, Dennis
  Jentsch, Hans Lamecker, Stefan Zachow, Frank Hertel, Jorge Goncalves, and
  Peter Gemmar.
\newblock Shape-aware surface reconstruction from sparse 3d point-clouds.
\newblock \emph{Medical image analysis}, 38:\penalty0 77--89, 2017.

\bibitem[Khallaghi et~al.(2015)Khallaghi, S{\'a}nchez, Rasoulian, Nouranian,
  Romagnoli, Abdi, Chang, Black, Goldenberg, Morris,
  et~al.]{khallaghi2015statistical}
Siavash Khallaghi, C~Antonio S{\'a}nchez, Abtin Rasoulian, Saman Nouranian,
  Cesare Romagnoli, Hamidreza Abdi, Silvia~D Chang, Peter~C Black, Larry
  Goldenberg, William~J Morris, et~al.
\newblock Statistical biomechanical surface registration: application to
  mr-trus fusion for prostate interventions.
\newblock \emph{IEEE transactions on medical imaging}, 34\penalty0
  (12):\penalty0 2535--2549, 2015.

\bibitem[Jud et~al.(2017)Jud, Giger, Sandk{\"u}hler, and
  Cattin]{jud2017localized}
Christoph Jud, Alina Giger, Robin Sandk{\"u}hler, and Philippe~C Cattin.
\newblock A localized statistical motion model as a reproducing kernel for
  non-rigid image registration.
\newblock In \emph{International Conference on Medical Image Computing and
  Computer-Assisted Intervention}, pages 261--269. Springer, 2017.

\bibitem[Wilms et~al.(2017)Wilms, Handels, and Ehrhardt]{wilms2017multi}
Matthias Wilms, Heinz Handels, and Jan Ehrhardt.
\newblock Multi-resolution multi-object statistical shape models based on the
  locality assumption.
\newblock \emph{Medical image analysis}, 38:\penalty0 17--29, 2017.

\bibitem[Wilms et~al.(2020)Wilms, Ehrhardt, and Forkert]{wilms2020kernelized}
Matthias Wilms, Jan Ehrhardt, and Nils~D Forkert.
\newblock A kernelized multi-level localization method for flexible shape
  modeling with few training data.
\newblock In \emph{Medical Image Computing and Computer Assisted
  Intervention--MICCAI 2020: 23rd International Conference, Lima, Peru, October
  4--8, 2020, Proceedings, Part IV 23}, pages 765--775. Springer, 2020.

\bibitem[Cerrolaza et~al.(2011)Cerrolaza, Villanueva, and
  Cabeza]{cerrolaza2011multi}
Juan~J Cerrolaza, Arantxa Villanueva, and Rafael Cabeza.
\newblock Multi-shape-hierarchical active shape models.
\newblock In \emph{Proceedings of the International Conference on Image
  Processing, Computer Vision, and Pattern Recognition (IPCV)}, page~1.
  Citeseer, 2011.

\bibitem[Zhang et~al.(2014)Zhang, Malcolm, Hislop-Jambrich, Thomas, and
  Nielsen]{zhang2014anatomical}
Ju~Zhang, Duane Malcolm, Jacqui Hislop-Jambrich, C~David~L Thomas, and Poul~MF
  Nielsen.
\newblock An anatomical region-based statistical shape model of the human
  femur.
\newblock \emph{Computer Methods in Biomechanics and Biomedical Engineering:
  Imaging \& Visualization}, 2\penalty0 (3):\penalty0 176--185, 2014.

\bibitem[Avants et~al.(2006)Avants, Schoenemann, and Gee]{avants2006lagrangian}
Brian~B Avants, P~Thomas Schoenemann, and James~C Gee.
\newblock Lagrangian frame diffeomorphic image registration: Morphometric
  comparison of human and chimpanzee cortex.
\newblock \emph{Medical image analysis}, 10\penalty0 (3):\penalty0 397--412,
  2006.

\bibitem[Ashburner and Ridgway(2013)]{ashburner2013symmetric}
John Ashburner and Gerard~R Ridgway.
\newblock Symmetric diffeomorphic modeling of longitudinal structural mri.
\newblock \emph{Frontiers in neuroscience}, 6:\penalty0 197, 2013.

\bibitem[Lui et~al.(2012)Lui, Wong, Zeng, Gu, Thompson, Chan, and
  Yau]{lui2012optimization}
Lok~Ming Lui, Tsz~Wai Wong, Wei Zeng, Xianfeng Gu, Paul~M Thompson, Tony~F
  Chan, and Shing-Tung Yau.
\newblock Optimization of surface registrations using beltrami holomorphic
  flow.
\newblock \emph{Journal of scientific computing}, 50:\penalty0 557--585, 2012.

\bibitem[Miller et~al.(2014)Miller, Younes, and
  Trouv{\'e}]{miller2014diffeomorphometry}
Michael~I Miller, Laurent Younes, and Alain Trouv{\'e}.
\newblock Diffeomorphometry and geodesic positioning systems for human anatomy.
\newblock \emph{Technology}, 2\penalty0 (01):\penalty0 36--43, 2014.

\bibitem[Sun et~al.(2022)Sun, Han, Kong, Tang, Yan, and Xie]{sun2022topology}
Shanlin Sun, Kun Han, Deying Kong, Hao Tang, Xiangyi Yan, and Xiaohui Xie.
\newblock Topology-preserving shape reconstruction and registration via neural
  diffeomorphic flow.
\newblock In \emph{Proceedings of the IEEE/CVF Conference on Computer Vision
  and Pattern Recognition}, pages 20845--20855, 2022.

\bibitem[Heimann and Meinzer(2009)]{heimann2009statistical}
Tobias Heimann and Hans-Peter Meinzer.
\newblock Statistical shape models for 3d medical image segmentation: a review.
\newblock \emph{Medical image analysis}, 13\penalty0 (4):\penalty0 543--563,
  2009.

\bibitem[Bohlender et~al.(2021)Bohlender, Oksuz, and
  Mukhopadhyay]{bohlender2021survey}
Simon Bohlender, Ilkay Oksuz, and Anirban Mukhopadhyay.
\newblock A survey on shape-constraint deep learning for medical image
  segmentation.
\newblock \emph{IEEE Reviews in Biomedical Engineering}, 2021.

\bibitem[El~Jurdi et~al.(2021)El~Jurdi, Petitjean, Honeine, Cheplygina, and
  Abdallah]{el2021high}
Rosana El~Jurdi, Caroline Petitjean, Paul Honeine, Veronika Cheplygina, and
  Fahed Abdallah.
\newblock High-level prior-based loss functions for medical image segmentation:
  A survey.
\newblock \emph{Computer Vision and Image Understanding}, 210:\penalty0 103248,
  2021.

\bibitem[Pizer et~al.(2013)Pizer, Jung, Goswami, Vicory, Zhao, Chaudhuri,
  Damon, Huckemann, and Marron]{pizer2013nested}
Stephen~M Pizer, Sungkyu Jung, Dibyendusekhar Goswami, Jared Vicory, Xiaojie
  Zhao, Ritwik Chaudhuri, James~N Damon, Stephan Huckemann, and JS~Marron.
\newblock Nested sphere statistics of skeletal models.
\newblock \emph{Innovations for shape analysis: Models and algorithms}, pages
  93--115, 2013.

\bibitem[Yushkevich et~al.(2019)Yushkevich, Aly, Wang, Xie, Gorman, Younes, and
  Pouch]{yushkevich2019diffeomorphic}
Paul~A Yushkevich, Ahmed Aly, Jiancong Wang, Long Xie, Robert~C Gorman, Laurent
  Younes, and Alison~M Pouch.
\newblock Diffeomorphic medial modeling.
\newblock In \emph{International Conference on Information Processing in
  Medical Imaging}, pages 208--220. Springer, 2019.

\bibitem[Cootes and Taylor(2001)]{cootes2001statistical}
Tim~F Cootes and Christopher~J Taylor.
\newblock Statistical models of appearance for medical image analysis and
  computer vision.
\newblock In \emph{Medical Imaging 2001: Image Processing}, volume 4322, pages
  236--248. SPIE, 2001.

\bibitem[Yenamandra et~al.(2021)Yenamandra, Tewari, Bernard, Seidel, Elgharib,
  Cremers, and Theobalt]{yenamandra2021i3dmm}
Tarun Yenamandra, Ayush Tewari, Florian Bernard, Hans-Peter Seidel, Mohamed
  Elgharib, Daniel Cremers, and Christian Theobalt.
\newblock i3dmm: Deep implicit 3d morphable model of human heads.
\newblock In \emph{Proceedings of the IEEE/CVF Conference on Computer Vision
  and Pattern Recognition}, pages 12803--12813, 2021.

\bibitem[Zheng et~al.(2017)Zheng, Li, and Szekely]{zheng2017statistical}
Guoyan Zheng, Shuo Li, and Gabor Szekely.
\newblock \emph{Statistical shape and deformation analysis: methods,
  implementation and applications}.
\newblock Academic Press, 2017.

\bibitem[Zhang and Fletcher(2013)]{zhang2013probabilistic}
Miaomiao Zhang and Tom Fletcher.
\newblock Probabilistic principal geodesic analysis.
\newblock \emph{Advances in neural information processing systems}, 26, 2013.

\bibitem[Yu et~al.(2021)Yu, Rao, Wang, Liu, Lu, and Zhou]{yu2021pointr}
Xumin Yu, Yongming Rao, Ziyi Wang, Zuyan Liu, Jiwen Lu, and Jie Zhou.
\newblock Pointr: Diverse point cloud completion with geometry-aware
  transformers.
\newblock In \emph{Proceedings of the IEEE/CVF international conference on
  computer vision}, pages 12498--12507, 2021.

\bibitem[Chibane et~al.(2020)Chibane, Alldieck, and
  Pons-Moll]{chibane2020implicit}
Julian Chibane, Thiemo Alldieck, and Gerard Pons-Moll.
\newblock Implicit functions in feature space for 3d shape reconstruction and
  completion.
\newblock In \emph{Proceedings of the IEEE/CVF conference on computer vision
  and pattern recognition}, pages 6970--6981, 2020.

\bibitem[Srivastava et~al.(2009)Srivastava, Samir, Joshi, and
  Daoudi]{srivastava2009elastic}
Anuj Srivastava, Chafik Samir, Shantanu~H Joshi, and Mohamed Daoudi.
\newblock Elastic shape models for face analysis using curvilinear coordinates.
\newblock \emph{Journal of Mathematical Imaging and Vision}, 33:\penalty0
  253--265, 2009.

\bibitem[Reuter et~al.(2009)Reuter, Wolter, Shenton, and
  Niethammer]{reuter2009laplace}
Martin Reuter, Franz-Erich Wolter, Martha Shenton, and Marc Niethammer.
\newblock Laplace--beltrami eigenvalues and topological features of
  eigenfunctions for statistical shape analysis.
\newblock \emph{Computer-Aided Design}, 41\penalty0 (10):\penalty0 739--755,
  2009.

\bibitem[Li et~al.(2022)Li, Huang, and Tian]{li2022comprehensive}
Menghan Li, Bin Huang, and Guohui Tian.
\newblock A comprehensive survey on 3d face recognition methods.
\newblock \emph{Engineering Applications of Artificial Intelligence},
  110:\penalty0 104669, 2022.

\bibitem[Egger et~al.(2020)Egger, Smith, Tewari, Wuhrer, Zollhoefer, Beeler,
  Bernard, Bolkart, Kortylewski, Romdhani, et~al.]{egger20203d}
Bernhard Egger, William~AP Smith, Ayush Tewari, Stefanie Wuhrer, Michael
  Zollhoefer, Thabo Beeler, Florian Bernard, Timo Bolkart, Adam Kortylewski,
  Sami Romdhani, et~al.
\newblock 3d morphable face models—past, present, and future.
\newblock \emph{ACM Transactions on Graphics (TOG)}, 39\penalty0 (5):\penalty0
  1--38, 2020.

\bibitem[Gerig et~al.(2018)Gerig, Morel-Forster, Blumer, Egger, Luthi,
  Sch{\"o}nborn, and Vetter]{gerig2018morphable}
Thomas Gerig, Andreas Morel-Forster, Clemens Blumer, Bernhard Egger, Marcel
  Luthi, Sandro Sch{\"o}nborn, and Thomas Vetter.
\newblock Morphable face models-an open framework.
\newblock In \emph{2018 13th IEEE International Conference on Automatic Face \&
  Gesture Recognition (FG 2018)}, pages 75--82. IEEE, 2018.

\bibitem[Genova et~al.(2018)Genova, Cole, Maschinot, Sarna, Vlasic, and
  Freeman]{genova2018unsupervised}
Kyle Genova, Forrester Cole, Aaron Maschinot, Aaron Sarna, Daniel Vlasic, and
  William~T Freeman.
\newblock Unsupervised training for 3d morphable model regression.
\newblock In \emph{Proceedings of the IEEE Conference on Computer Vision and
  Pattern Recognition}, pages 8377--8386, 2018.

\bibitem[Wood et~al.(2021)Wood, Baltru{\v{s}}aitis, Hewitt, Dziadzio, Cashman,
  and Shotton]{wood2021fake}
Erroll Wood, Tadas Baltru{\v{s}}aitis, Charlie Hewitt, Sebastian Dziadzio,
  Thomas~J Cashman, and Jamie Shotton.
\newblock Fake it till you make it: face analysis in the wild using synthetic
  data alone.
\newblock In \emph{Proceedings of the IEEE/CVF international conference on
  computer vision}, pages 3681--3691, 2021.

\bibitem[Grassal et~al.(2022)Grassal, Prinzler, Leistner, Rother, Nie{\ss}ner,
  and Thies]{grassal2022neural}
Philip-William Grassal, Malte Prinzler, Titus Leistner, Carsten Rother,
  Matthias Nie{\ss}ner, and Justus Thies.
\newblock Neural head avatars from monocular rgb videos.
\newblock In \emph{Proceedings of the IEEE/CVF Conference on Computer Vision
  and Pattern Recognition}, pages 18653--18664, 2022.

\bibitem[Zheng et~al.(2022)Zheng, Abrevaya, B{\"u}hler, Chen, Black, and
  Hilliges]{zheng2022avatar}
Yufeng Zheng, Victoria~Fern{\'a}ndez Abrevaya, Marcel~C B{\"u}hler, Xu~Chen,
  Michael~J Black, and Otmar Hilliges.
\newblock Im avatar: Implicit morphable head avatars from videos.
\newblock In \emph{Proceedings of the IEEE/CVF Conference on Computer Vision
  and Pattern Recognition}, pages 13545--13555, 2022.

\bibitem[Adams et~al.(2004)Adams, Rohlf, and Slice]{adams2004geometric}
Dean~C Adams, F~James Rohlf, and Dennis~E Slice.
\newblock Geometric morphometrics: ten years of progress following the
  ‘revolution’.
\newblock \emph{Italian journal of zoology}, 71\penalty0 (1):\penalty0 5--16,
  2004.

\bibitem[Cates et~al.(2017)Cates, Nevell, Prajapati, Nelon, Chang, Randolph,
  Wood, Keller, and Whitaker]{cates2017shape}
Joshua Cates, Lisa Nevell, Suresh~I Prajapati, Laura~D Nelon, Jerry~Y Chang,
  Matthew~E Randolph, Bernard Wood, Charles Keller, and Ross~T Whitaker.
\newblock Shape analysis of the basioccipital bone in pax7-deficient mice.
\newblock \emph{Scientific reports}, 7\penalty0 (1):\penalty0 17955, 2017.

\bibitem[Baka et~al.(2010)Baka, de~Bruijne, Reiber, Niessen, and
  Lelieveldt]{baka2010confidence}
Nora Baka, Marleen de~Bruijne, Johan~HC Reiber, Wiro Niessen, and Boudewijn~PF
  Lelieveldt.
\newblock Confidence of model based shape reconstruction from sparse data.
\newblock In \emph{2010 IEEE International Symposium on Biomedical Imaging:
  From Nano to Macro}, pages 1077--1080. IEEE, 2010.

\bibitem[Salhi et~al.(2020)Salhi, Burdin, Boutillon, Brochard, Mutsvangwa, and
  Borotikar]{salhi2020statistical}
Asma Salhi, Val{\'e}rie Burdin, Arnaud Boutillon, Sylvain Brochard, Tinashe
  Mutsvangwa, and Bhushan Borotikar.
\newblock Statistical shape modeling approach to predict missing scapular bone.
\newblock \emph{Annals of biomedical engineering}, 48\penalty0 (1):\penalty0
  367--379, 2020.

\bibitem[Zhang and Besier(2017)]{zhang2017accuracy}
Ju~Zhang and Thor~F Besier.
\newblock Accuracy of femur reconstruction from sparse geometric data using a
  statistical shape model.
\newblock \emph{Computer methods in biomechanics and biomedical engineering},
  20\penalty0 (5):\penalty0 566--576, 2017.

\bibitem[Gower(1975)]{gower1975generalized}
John~C Gower.
\newblock Generalized procrustes analysis.
\newblock \emph{Psychometrika}, 40\penalty0 (1):\penalty0 33--51, 1975.

\bibitem[L{\"u}thi et~al.(2017)L{\"u}thi, Gerig, Jud, and
  Vetter]{luthi2017gaussian}
Marcel L{\"u}thi, Thomas Gerig, Christoph Jud, and Thomas Vetter.
\newblock Gaussian process morphable models.
\newblock \emph{IEEE transactions on pattern analysis and machine
  intelligence}, 40\penalty0 (8):\penalty0 1860--1873, 2017.

\bibitem[Drineas et~al.(2005)Drineas, Mahoney, and
  Cristianini]{drineas2005nystrom}
Petros Drineas, Michael~W Mahoney, and Nello Cristianini.
\newblock On the nystr{\"o}m method for approximating a gram matrix for
  improved kernel-based learning.
\newblock \emph{journal of machine learning research}, 6\penalty0 (12), 2005.

\bibitem[D{\"o}lz et~al.(2019)D{\"o}lz, Gerig, L{\"u}thi, Harbrecht, and
  Vetter]{dolz2019error}
J{\"u}rgen D{\"o}lz, Thomas Gerig, Marcel L{\"u}thi, Helmut Harbrecht, and
  Thomas Vetter.
\newblock Error-controlled model approximation for gaussian process morphable
  models.
\newblock \emph{Journal of Mathematical Imaging and Vision}, 61:\penalty0
  443--457, 2019.

\bibitem[Tipping and Bishop(1999)]{tipping1999probabilistic}
Michael~E Tipping and Christopher~M Bishop.
\newblock Probabilistic principal component analysis.
\newblock \emph{Journal of the Royal Statistical Society: Series B (Statistical
  Methodology)}, 61\penalty0 (3):\penalty0 611--622, 1999.

\bibitem[Talts et~al.(2018)Talts, Betancourt, Simpson, Vehtari, and
  Gelman]{talts2018validating}
Sean Talts, Michael Betancourt, Daniel Simpson, Aki Vehtari, and Andrew Gelman.
\newblock Validating bayesian inference algorithms with simulation-based
  calibration.
\newblock \emph{arXiv preprint arXiv:1804.06788}, 2018.

\bibitem[S{\"a}rkk{\"a}(2011)]{sarkka2011linear}
Simo S{\"a}rkk{\"a}.
\newblock Linear operators and stochastic partial differential equations in
  gaussian process regression.
\newblock In \emph{Artificial Neural Networks and Machine Learning--ICANN 2011:
  21st International Conference on Artificial Neural Networks, Espoo, Finland,
  June 14-17, 2011, Proceedings, Part II 21}, pages 151--158. Springer, 2011.

\bibitem[Besl and McKay(1992)]{besl1992method}
Paul~J Besl and Neil~D McKay.
\newblock Method for registration of 3-d shapes.
\newblock In \emph{Sensor fusion IV: control paradigms and data structures},
  volume 1611, pages 586--606. International Society for Optics and Photonics,
  1992.

\bibitem[Zhang(1994)]{zhang1994iterative}
Zhengyou Zhang.
\newblock Iterative point matching for registration of free-form curves and
  surfaces.
\newblock \emph{International journal of computer vision}, 13\penalty0
  (2):\penalty0 119--152, 1994.

\bibitem[Allen et~al.(2003)Allen, Curless, and Popovi{\'c}]{allen2003space}
Brett Allen, Brian Curless, and Zoran Popovi{\'c}.
\newblock The space of human body shapes: reconstruction and parameterization
  from range scans.
\newblock \emph{ACM transactions on graphics (TOG)}, 22\penalty0 (3):\penalty0
  587--594, 2003.

\bibitem[Amberg et~al.(2007)Amberg, Romdhani, and Vetter]{amberg2007optimal}
Brian Amberg, Sami Romdhani, and Thomas Vetter.
\newblock Optimal step nonrigid icp algorithms for surface registration.
\newblock In \emph{2007 IEEE conference on computer vision and pattern
  recognition}, pages 1--8. IEEE, 2007.

\bibitem[Robert et~al.(2010)Robert, Casella, Robert, and
  Casella]{robert2010metropolis}
Christian Robert, George Casella, Christian~P Robert, and George Casella.
\newblock Metropolis--hastings algorithms.
\newblock \emph{Introducing Monte Carlo Methods with R}, pages 167--197, 2010.

\bibitem[Sch{\"o}nborn et~al.(2017)Sch{\"o}nborn, Egger, Morel-Forster, and
  Vetter]{schonborn2017markov}
Sandro Sch{\"o}nborn, Bernhard Egger, Andreas Morel-Forster, and Thomas Vetter.
\newblock Markov chain monte carlo for automated face image analysis.
\newblock \emph{International Journal of Computer Vision}, 123\penalty0
  (2):\penalty0 160--183, 2017.

\bibitem[Madsen et~al.(2020)Madsen, Morel-Forster, Kahr, Rahbani, Vetter, and
  L{\"u}thi]{madsen2020closest}
Dennis Madsen, Andreas Morel-Forster, Patrick Kahr, Dana Rahbani, Thomas
  Vetter, and Marcel L{\"u}thi.
\newblock A closest point proposal for mcmc-based probabilistic surface
  registration.
\newblock In \emph{Computer Vision--ECCV 2020: 16th European Conference,
  Glasgow, UK, August 23--28, 2020, Proceedings, Part XVII 16}, pages 281--296.
  Springer, 2020.

\bibitem[Rezende and Mohamed(2015)]{rezende2015variational}
Danilo Rezende and Shakir Mohamed.
\newblock Variational inference with normalizing flows.
\newblock In \emph{International conference on machine learning}, pages
  1530--1538. PMLR, 2015.

\bibitem[Papamakarios et~al.(2021)Papamakarios, Nalisnick, Rezende, Mohamed,
  and Lakshminarayanan]{papamakarios2021normalizing}
George Papamakarios, Eric Nalisnick, Danilo~Jimenez Rezende, Shakir Mohamed,
  and Balaji Lakshminarayanan.
\newblock Normalizing flows for probabilistic modeling and inference.
\newblock \emph{Journal of Machine Learning Research}, 22\penalty0
  (57):\penalty0 1--64, 2021.

\bibitem[Perugachi-Diaz et~al.(2021)Perugachi-Diaz, Tomczak, and
  Bhulai]{perugachi2021invertible}
Yura Perugachi-Diaz, Jakub Tomczak, and Sandjai Bhulai.
\newblock Invertible densenets with concatenated lipswish.
\newblock \emph{Advances in Neural Information Processing Systems},
  34:\penalty0 17246--17257, 2021.

\bibitem[Van~Laarhoven et~al.(1987)Van~Laarhoven, Aarts, van Laarhoven, and
  Aarts]{van1987simulated}
Peter~JM Van~Laarhoven, Emile~HL Aarts, Peter~JM van Laarhoven, and Emile~HL
  Aarts.
\newblock \emph{Simulated annealing}.
\newblock Springer, 1987.

\bibitem[Akhbari et~al.(2019)Akhbari, Moore, Laidlaw, Weiss, Akelman, Wolfe,
  and Crisco]{akhbari2019predicting}
Bardiya Akhbari, Douglas~C Moore, David~H Laidlaw, Arnold-Peter~C Weiss, Edward
  Akelman, Scott~W Wolfe, and Joseph~J Crisco.
\newblock Predicting carpal bone kinematics using an expanded digital database
  of wrist carpal bone anatomy and kinematics.
\newblock \emph{Journal of Orthopaedic Research{\textregistered}}, 37\penalty0
  (12):\penalty0 2661--2670, 2019.

\end{thebibliography}

\clearpage
\appendix

\section{Likelihood}
\label{sec:appendixLikelihood}

Given the target $\Gamma_{\tau}$ observed on $Z \subseteq \Omega$ and using a discretization $\{x_i\}_i$ of the domain $Z$ the likelihood is defined as

\begin{equation}
\label{eq:fullFormulationLikelihood}
p^X(\Gamma_{\tau} | \theta) = \prod_{x_i \in Z^N} \mathcal{N}\bigl(CP(x, \Gamma_\tau); x, \sigma^2\bigr),
\end{equation}

where the function $CP(x,\Gamma_\tau)$ returns the corresponding point of $x$ on $\Gamma_\tau$, and $x = \Gamma[\theta](x_i)$ relies on the shape model $\GP(\mu^X, k^X)$ leading to $\Gamma[\theta](x) = R_{\boldsymbol r}(x+u^X[\boldsymbol\alpha](x))+\boldsymbol t$.

Every possible domain that is chosen for the rigid alignment objective in \cref{eq:gpaAlignmentOperator} corresponds to an individual shape model, which in turn induces a posterior distribution. As such given a domain $X$, we identify the shape model as $\GP(\mu^X, k^X)$, the likelihood as $p^X(\Gamma_{\tau} | \theta)$ and the posterior as $p^X\!(\theta | \Gamma_\tau)$ or shortened to $p^X$.

\section{Details - femur reconstruction}
\label{sec:detailedResults}
We will discuss more details of the femur reconstruction experiment from \cref{sec:fullInference}. While we list more details this is not exhaustive. For more information we refer the reader to the available code \footnote{\url{https://github.com/unibas-gravis/posterior-shape-models-revisited}}.

It is important to keep in mind when looking at the squared distance metric that the reconstructions are well fitting the observed domain. The average squared distance on $X$ to the closest point on the target surface is generally below $0.5$ mm$^2$ with few exceptions. The reported distance in the Figures and Tables remains high as we list the distance to the true corresponding point on the target. The correspondence is unknown during optimization. Instead the heuristic of the closest point on the target surface is employed. But we do not report those numbers as we are focusing on reconstructive quality.

For nICP and MH we use an unbiased pose initialization. We follow \cref{eq:gpaAlignmentOperator} for the domain $X$ and add Gaussian noise on translation $2$ mm std and rotation $0.1$ rad std to get a unbiased noisy pose initialization. In general, one can always use a rigid alignment first to adequately align the initialization, therefore this has little to no effect on the final reconstruction quality if convergence is successful. The normalizing flow is initialized with a similarly accurate initial distribution to avoid pathological solutions.

For the likelihood we assume i.i.d. normal distributed noise with $1$ mm std. We run the inference algorithms with a common computational budget. That means less than a minute for nICP, $30$ min for MH and $1.5$ h on the GPU for normalizing flows.

\subsection{nICP reconstruction}
\label{sec:detailedResultsNICP}

The details are listed in \cref{tab:expTableIcp}. We list the expected distance for nICP reconstruction separated into different regions and for additional alignment methods. \Cref{fig:icpPost} lists these values only over the full domain $\Omega$. In \Cref{tab:expTableIcp} the metrics are separated into two additional regions, the observed region $X$ and the predicted region $Z=\Omega-X$.

\Cref{tab:expTableIcp} lists additional variants where we use our projection while using the ground-truth-$X$ to ablate that our method only uses an estimate of $X$. The target-specific $p^X$ we compare to uses the ground-truth-$X$ for $p^X$. This is idealized and does not reflect practical limitations and as such $p^X$ is a strong baseline. We additionally test landmark $L\subset X$ aligned model induced posterior distribution as well. For more details see \cref{sec:landmarkDiscussionAppendix}.

All the methods that are aligned on $X$ have a similar performance. The linear approximation of rotation seems to benefit the reconstruction for the femur data. But this is not something that should be expected to generalize.

Compared to the MH reconstruction, one can see a noticeably less accurate reconstruction with observed ratios higher than $45\%$. nICP with its gradient descent like approach is not ideal for reconstructions. The gradients can quickly converge to a local optima with a bad correspondence estimation. Generally, the method struggles to escape these local minima.

We want to emphasize again that the shown distance does not portray the distance to the surface. What we show is the distance to the corresponding point which can be seen as the error in correspondence. This metric is always higher than the distance to the closest point on the target surface.

\begin{figure}
    \centering
\includegraphics[width=0.6\linewidth]{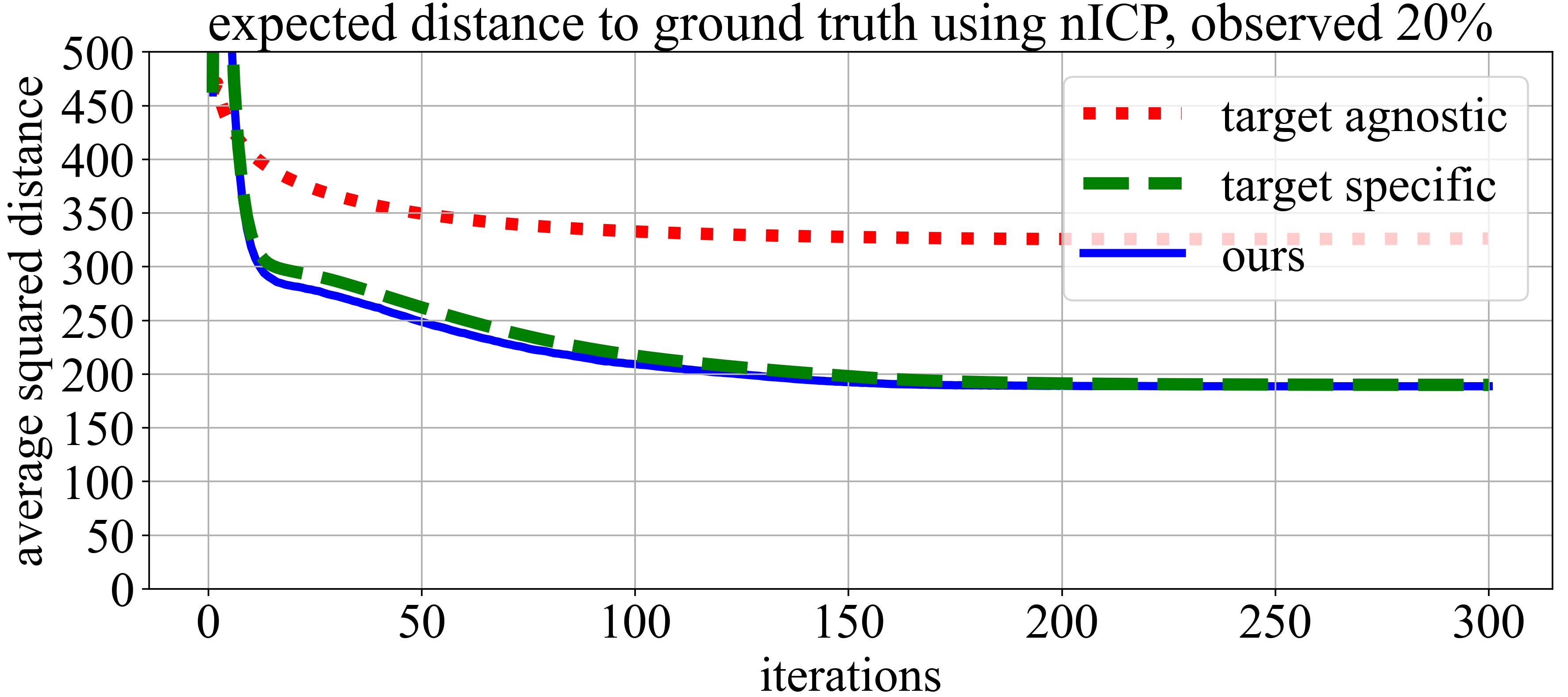}
    \caption{Expected distance of the nICP reconstruction experiment. The $x$-axis shows how much of the femur was observed. The $y$ lists the distance to the ground truth femur.}
    \label{fig:icpConvPost}
\end{figure}

\Cref{fig:icpConvPost} lists the difference in convergence speed between the shape models for the nICP reconstruction experiment. For observing 20\% of the target the nICP reconstructive quality is shown with the required number of optimization iterations. In the beginning target-specific models get worse for one iteration unlike the target-agnostic ones. This is mainly because of the errors in pose initialization lead to early correspondence mistakes during convergence. The target-agnostic models ignore these mistakes because of the high dependence of pose and shape that requires numerous iterations to partially resolve. Overall, the target-specific models converge in fewer iterations to a reasonable reconstruction, while the target-agnostic models do not converge to a similarly good reconstruction.

Overall, reconstructions using nICP are fairly good for small observation ratios $\leq 30\%$, also considering the low computational costs. It can serve as a valid initialization for other methods, but as performance suffers when more of the target is observed and as the method does not provide a distribution of reconstructions, we do not recommend to use nICP for general applications.

\begin{table*}
\begin{center}
\setlength\tabcolsep{4pt}
\resizebox{\textwidth}{!}{%
\begin{tabular}{|l|c|c|c|c|c|c|c|c|c|c|c|c|c|c|c|}
\hline
ratio observed & \multicolumn{3}{c}{20\%} \vline & \multicolumn{3}{c}{30\%} \vline & \multicolumn{3}{c}{45\%} \vline & \multicolumn{3}{c}{60\%} \vline & \multicolumn{3}{c}{80\%} \vline \\ 
\hline
region $~~(Z = \Omega - X)$ & $X$ & $Z$ & $\Omega$ & $X$ & $Z$ & $\Omega$ & $X$ & $Z$ & $\Omega$ & $X$ & $Z$ & $\Omega$ & $X$ & $Z$ & $\Omega$ \\ 
\hline \hline
target-agnostic $p^\Omega$ & 3.5&554&368&9.7&500&305&21&447&249&31&391&198&46&313&139 \\
\rom{1}) $p^L$ & 2.1&315&209&4.6&266&163&13&248&139&26&259&134&56&273&131 \\
\rom{2}) $p^L$ & 2.1&314&209&4.7&269&164&14&262&147&29&294&151&82&582&253 \\
\rom{3}) $p^{\tilde{L}}$ & \tb{1.9}&309&205&4.6&265&162&14&261&146&29&298&153&89&632&274 \\
target-specific $p^X$ & \tb{1.9}&309&205&\tb{3.9}&254&155&9.3&211&117&14&169&86&23&138&63 \\ \hline
ours $p^\mathcal{P}$ & \tb{1.9}&\tb{304}&\tb{202}&\tb{3.9}&252&\tb{153}&9.3&206&115&14&167&85&23&139&63 \\
\rom{4}) ours GT-$X$ & \tb{1.9}&\tb{304}&\tb{202}&\tb{3.9}&\tb{251}&\tb{153}&\tb{8.9}&\tb{200}&\tb{111}&\tb{13}&\tb{159}&\tb{81}&\tb{21}&\tb{127}&\tb{58} \\
\hline
\end{tabular}}
\end{center}
\caption{Extended version of \cref{tab:expTableShort} including additionally tested models for the nICP method. Listed is the average of the squared distance to the ground truth target using the ground truth correspondence separated into three regions: observed domain $X$, predicted domain $Z$ and full domain $\Omega$. In reconstruction tasks the error in $Z$ is of great interest. \rom{1}) the target is aligned to the mean on $X$. \rom{2}) the target is aligned to the mean on $L$. \rom{3}) the landmarks for alignment of all shapes are observed with Gaussian noise $\sigma=1$. \rom{4}) the ground-truth-$X$ is used for the projection. Lowest prediction distance in bold.}
\label{tab:expTableIcp}
\end{table*}

\subsection{MH reconstruction}
\label{sec:detailedResultsMH}

\begin{table*}
\begin{center}
\setlength\tabcolsep{4pt}
\resizebox{\textwidth}{!}{%
\begin{tabular}{|l|l|c|c|c|c|c|c|c|c|c|c|c|c|c|c|c|}
\hline
\multicolumn{2}{|l}{ratio observed} \vline & \multicolumn{3}{c}{20\%} \vline & \multicolumn{3}{c}{30\%} \vline & \multicolumn{3}{c}{45\%} \vline & \multicolumn{3}{c}{60\%} \vline & \multicolumn{3}{c}{80\%} \vline \\ 
\hline
\multicolumn{2}{|l}{region $~~(Z = \Omega - X)$} \vline & $X$ & $Z$ & $\Omega$ & $X$ & $Z$ & $\Omega$ & $X$ & $Z$ & $\Omega$ & $X$ & $Z$ & $\Omega$ & $X$ & $Z$ & $\Omega$ \\ 
\hline \hline

\multirow{2}{*}{$p^\Omega$} & MSE & 1.0&479&318&3.7&411&249 & 7.1 & 220 & 122 & 11 & 179 & 90 & 5.8 & 54 & 23 \\ 
& mean var & .52 & 63 & 42 & .69 & 42 & 26 & 1.3 & 30 & 17 & 1.9 & 27 & 13 & 3.8 & 30 & 13 \\\hline 
\multirow{2}{*}{\rom{1}) $p^L$} & MSE & \tb{.48}&314&208&1.2&231&140 & 2.8 & 151 & 83 & 3.6 & 81 & 40 & 6.4 & 52 & 23 \\ 
& mean var & .59 & 107 & 71 & 1.0 & 79 & 48 & 1.3 & 41 & 23 & 1.7 & 31 & 15 & 2.5 & 20 & 8.8 \\ \hline
\multirow{2}{*}{\rom{2}) $p^L$} & MSE & 2.8&395&262&7.3&331&203& 27 & 376 & 216 & 58 & 474 & 254 & 142 & 507 & 270 \\ 
& mean var & .59 & 108 & 72 & 1.1 & 81 & 50 & 1.5 & 45 & 25 & 1.9 & 30 & 15 & 3.2 & 24 & 10 \\ \hline
\multirow{2}{*}{\rom{3}) $p^{\tilde{L}}$} & MSE & 4.4&456&303&9.3&431&264&32 & 476 & 272 & 57 & 484 & 258 & 174 & 721 & 367 \\
& mean var & .58 & 106 & 70 & .98 & 76 & 46 & 1.4 & 43 & 24 & 1.8 & 29 & 15 & 3.1 & 22 & 10 \\ \hline
\multirow{2}{*}{$p^X$} & MSE & .49&316&209&\tb{.93}&235&142& 4.4 & 166 & 92 & 3.5 & 85 & 42 & 5.0 & \tb{48} & \tb{20} \\
& mean var & .59 & 110 & 73 & .97 & 78 & 48 & 1.4 & 44 & 24 & 1.4 & 26 & 13 & 1.4 & 12 & 5.0 \\ \hline \hline
\multirow{2}{*}{ours $p^\mathcal{P}$} & MSE & .52&304&201&1.1&236&143&5.7 & 185 & 103 & 3.7 & \tb{75} & \tb{37} & 6.0 & 55 & 23 \\
& mean var & .60 & 109 & 73 & 1.1 & 78 & 48 & 1.3 & 42 & 23 & 1.4 & 26 & 13 & 1.7 & 14 & 6.0 \\ \hline
\multirow{1}{*}{\rom{4}) ours} & MSE & .50&\tb{293}&\tb{194}&1.0&\tb{223}&\tb{135}& \tb{1.2} & \tb{132} & \tb{72} & \tb{2.8} & 79 & \tb{37} & \tb{3.5} & 64 & 25 \\
\multirow{1}{*}{GT-$X$}& mean var & .60 & 107 & 72 & 1.0 & 79 & 48 & 1.4 & 43 & 24 & 1.4 & 26 & 13 & 1.8 & 16 & 6.9 \\
\hline
\end{tabular}}
\end{center}
\caption{Extended version of \cref{tab:expTableShort} including additionally tested models for the MH method. Listed are the mean squared error (MSE) to the ground truth target using the correct correspondence, and the mean of the predicted uncertainty. The metrics are listed for three regions separately: observed domain $X$, predicted domain $Z$ and full domain $\Omega$. In reconstruction tasks the error in $Z$ is of great interest. \rom{1}) the target is aligned to the mean on $X$. \rom{2}) the target is aligned to the mean on $L$. \rom{3}) the landmarks for alignment of all shapes are observed with Gaussian noise $\sigma=1$. \rom{4}) the ground truth $X$ is used for the projection. Lowest prediction distance in bold.}
\label{tab:expTable}
\end{table*}

The details are listed in \cref{tab:expTable}. This includes the expected distance and predicted variance in different regions. \Cref{fig:mhPost} lists these values only over the full domain $\Omega$. In \Cref{tab:expTable} the metrics are separated into two additional regions, the observed region $X$ and the predicted region $\Omega-X$.

The table includes four additional models. Three are landmark-aligned models. For a discussion see \cref{sec:landmarkDiscussionAppendix}. The landmarks are four distinct points in the distal part of the femur. As such $L \subset X$ for all the tested amounts of observed domain. When observing only $10\%$ of the target the landmarks are close to the border of the observed region. One can argue that the identification of the landmarks could be flawed in practice.

The three different landmark models differ in the alignment of the target and in the precision of the landmarks. The partial target can be aligned to model mean $\mu^L$ on $X$ or $L$. We test both alignments. So far we assumed perfectly identified landmarks. The third landmark model includes noise on the landmarks. We use Gaussian noise with $\sigma=1$ to all landmark observations, both in training and in the target. This can often happen in practice where there is ambiguity in expert labelling or automatic landmark identification. The alignment in \cref{eq:gpaAlignmentOperator} is robust to a big number of noisy observations. But often when using manually identified landmarks only a handful are identified.

The results show that aligning the target only on $L$ performs significantly worse than alignment on $X$ for landmark-aligned models. Additionally, the noise in the landmarks is a significant detriment to the reconstruction quality. As such, for this dataset the landmark-aligned models have competitive performance only under the most ideal conditions.

Our projected model resulting in $p^\mathcal{P}$ is also listed in two variants. For one we estimate $X$ iteratively during the Markov chain. In \Cref{sec:fullInference} we use this version as the projected model. The other variant uses the correct $X$ from start to finish. This can be seen as an ablation study. It is to demonstrate that our alignment method does not require iterative updates to be beneficial. The results show very similar performance between using the ground truth $X$ and updating an estimate of it.

The details are listed in \cref{tab:expTable}. This includes the expected distance and predicted variance in different regions. \Cref{fig:mhPost} lists these values only over the full domain $\Omega$. In \Cref{tab:expTable} the metrics are separated into two additional regions, the observed region $X$ and the predicted region $\Omega-X$.

\subsection{Normalizing flows}

\begin{table*}
\begin{center}
\setlength\tabcolsep{4pt}
\resizebox{\textwidth}{!}{%
\begin{tabular}{|l|l|c|c|c|c|c|c|c|c|c|c|c|c|c|c|c|}
\hline
\multicolumn{2}{|l}{ratio observed} \vline & \multicolumn{3}{c}{20\%} \vline & \multicolumn{3}{c}{30\%} \vline & \multicolumn{3}{c}{45\%} \vline & \multicolumn{3}{c}{60\%} \vline & \multicolumn{3}{c}{80\%} \vline \\ 
\hline
\multicolumn{2}{|l}{region $~~(Z = \Omega - X)$} \vline & $X$ & $Z$ & $\Omega$ & $X$ & $Z$ & $\Omega$ & $X$ & $Z$ & $\Omega$ & $X$ & $Z$ & $\Omega$ & $X$ & $Z$ & $\Omega$ \\ 
\hline \hline
\multirow{2}{*}{$p^\Omega$} & MSE & 1.0&316&209&1.7&243&147&2.3&133&72&3.2&97&47&3.7&67&25 \\
& mean var & 1.2&158&105&1.7&129&79&2.9&91&50&5.3&88&44&9.0&89&36 \\ \hline
\multirow{2}{*}{$p^X$} & MSE & 1.0&307&203&1.8&235&143&2.3&132&72&3.2&96&46&3.7&65&25 \\
& mean var & 1.3&225&149&1.7&154&93&2.6&98&54&4.7&91&45&8.1&89&36 \\ \hline
\multirow{2}{*}{ours $p^\mathcal{P}$} & MSE & \tb{.88}&\tb{288}&\tb{190}&\tb{1.5}&\tb{213}&\tb{129}&\tb{2.2}&\tb{124}&\tb{67}&\tb{2.6}&\tb{82}&\tb{39}&\tb{2.8}&\tb{57}&\tb{21} \\
& mean var & 1.3&224&148&1.7&153&93&2.7&100&55&4.9&96&47&8.6&93&37.5 \\ \hline
\end{tabular}}
\end{center}
\caption{Extended version of \cref{tab:expTableShort} for the normalizing flow method. Listed are the mean squared error (MSE) to the ground truth target using the correct correspondence, and the mean of the predicted uncertainty. The metrics are listed for three regions separately: observed domain $X$, predicted domain $Z$ and full domain $\Omega$. In reconstruction tasks the error in $\Omega$ or $Z$ is of great interest. We list fewer methods due to the computational costs. Lowest prediction distance in bold.}
\label{tab:detailedResultsNF}
\end{table*}

Due to the computational cost of normalizing flows we did not investigate the performance of the various other models, instead we limited the experiment to the important target-agnostic, target-specific and our projection model prior.

We use 8 layers of iResBlocks \cite{perugachi2021invertible} which provide a class of powerful bijections. During training we use a batch size of $8192$ parameter samples that are mapped through the flow. For each of these samples the posterior $p(\theta | \Gamma_\tau)$ is evaluated. Stacking more layers at the cost of reducing the batch size did not yield better performance. We train for $6000$ iterations which takes on average $1.5$ h using a A100.

normalizing flows are a modern deep-learning approach using powerful bijections to map a simple distribution into the powerful one. In addition, we also allocate more computational resources to training the method. Overall, it is better situated to overcome the bad conditioning of $p(\theta | \Gamma_\tau)$ due to the target-agnostic priors. Nevertheless, target-agnostic model priors still negatively affect the reconstructive quality as seen in \cref{tab:detailedResultsNF}. This time the reconstruction error is fairly similar between all models, but the variance of the target-specific models is better estimated. The target-agnostic posteriors seem to use up the available capacity of normalizing flows much more quickly due the highly interdependent pose and shape, seen with the much lower variance suggesting a slight concentration around the fairly successfully recovered MAPs.

\section{Landmark alignment}
\label{sec:landmarkDiscussionAppendix}

A common approach for training and target alignment is the use of easily identifiable landmarks. Given $X \subseteq \Omega$ the landmarks $L \subset X$ are in the observed region. An alignment on $L$ is often seen instead of an alignment to the complete observed domain. In this Section we discuss the behavior of landmark-aligned models $\GP(\mu^L, k^L)$ and their resulting posteriors $p^L\!(\theta | \Gamma_\tau)$.

Conceptually, there is no difference in handling the alignment to an arbitrary $X$ or $L$. Our projection alignment can be used to calculate the landmark-aligned model directly from $\GP(\mu^\Omega, k^\Omega)$. Still, at least three landmarks in $L$ are required for rotational alignment. When only aligning over translations, then a single landmark would give rise to a model which is aligned to a single point. That point is then a stationary point in the model. If the pose is fixed and not optimized, then the target should be rigidly aligned to a landmark model $\mu^L$ only on $L$ instead of the observed $X$ to avoid a different data distribution. However, if pose is also optimized then it is still recommended to align to the target on the entire $X$. This leads to a better starting point for the inference algorithm. The shape and pose remain interdependent posing more difficulties during inference. If the landmarks can be identified accurately, landmark-aligned models generally outperform target-agnostic models.

We recommend to not exclusively rely on landmarks but on as many target observations as possible. Noise on a few landmarks have a much stronger influence on the rigid alignment compared to an entire domain $X$. In other words, we recommend to align the data and target on $X$. Aligning on $L$ leads to good results if the pose is fixed and the landmark are accurately identified it is preferable to align a model on $X$ instead of a subset $L \subset X$. To showcase the performance of landmark-aligned models we also use them in the femur MH reconstruction experiment. \Cref{sec:detailedResults} lists more details.

Overall landmark aligned models are susceptible to noise to the low number of points to estimate the rigid alignment. Using noisy landmarks impacts performance negatively. This becomes highly noticeable, when the landmarks are positioned near the end of the femur at the distal part. Due to this positioning the negative impact is more strongly noticed when observing larger parts of the shape. Also when using posterior shape models one should align the target to the model mean on the domain $L$, not on the entire $X$. Interestingly, during the MH experiments aligning on $X$ performed better in many cases. Nevertheless, we recommend to use $p^X$ due its performance still being superior compared to $p^L$ even in a idealized setting with no noise on the landmarks. This can have exceptions if $X$ were difficult to estimate but there are easily identifiable $L$.

\section{Further experiments}

In this Section we describe additional experiments that investigate the rotation approximation error during projection for a synthetic shape and complex face shapes, using posterior shape models for reconstructions, and using a scaphoid dataset that verifies our findings on a completely different bone structure.

\subsection{Approximation error of realignment}
\label{sec:hingeAppendix}

\begin{figure}[!t]%
    \centering
\includegraphics[width=0.4\linewidth]{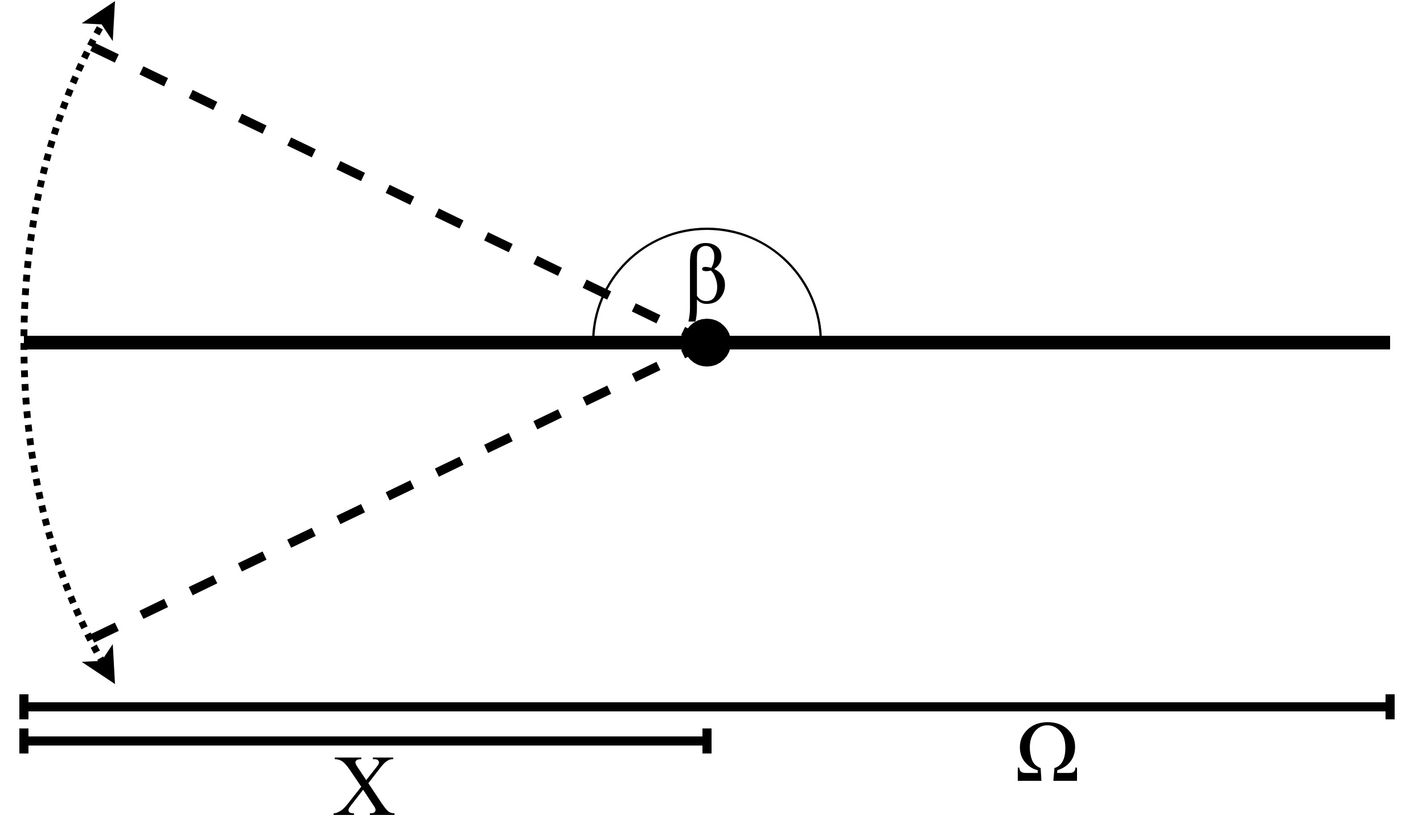}%
    \caption{The parametric rotation model is a model over shapes determined by the angle $\beta$. The left arm $X$ moves when the angle $\beta$ changes.}%
    \label{fig:hingeSetup}%
\end{figure}%

\begin{figure*}[!t]%
    \centering
\includegraphics[width=\linewidth]{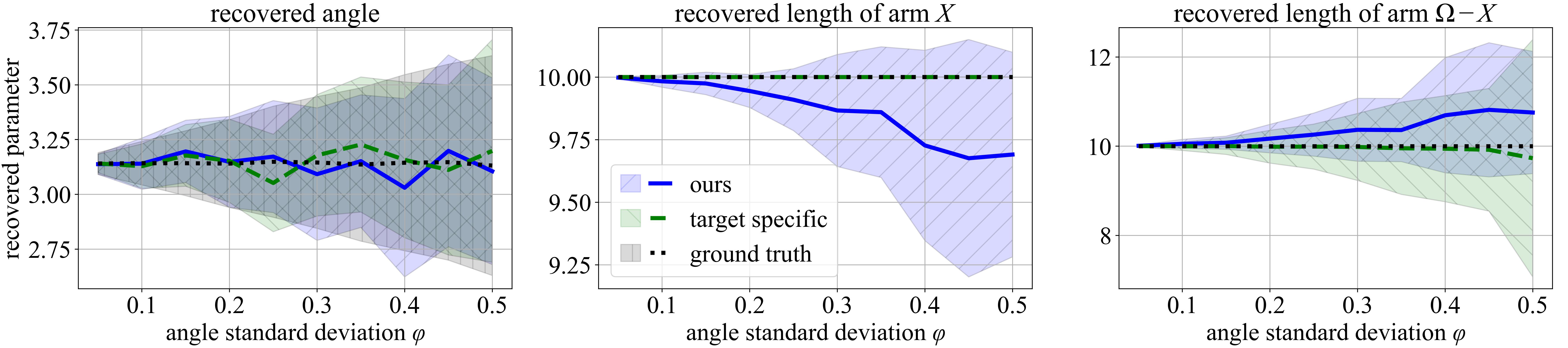}%
    \caption{The left shows the recovered distribution of the angle $\beta$. The middle and right show the recovered parameter for the arms of the hinge. One standard deviation is shown in the shaded area. Our method can imitate the behavior of a target-specific model adequately. Our method slightly worsens the arm shape distribution in the predicted domain (right) and introduces an error on the observed arm (middle).}%
    \label{fig:partialResHinge}%
\end{figure*}%

\Cref{sec:alignmentRot} lists the error of the approximation scaling with the sinus of the rotation. For small angles, i.e. $\le 15 \deg$, the approximation error is still fairly small. For most biological shapes rotations of parts that exceed such angles are rare. To investigate the error raising from larger rotations we define a minimal example with parametric rotation.

The true shape distribution is a simple mesh consisting of two equally long arms with a random angle $\beta \sim \mathcal{N}(\pi, \varphi)$ in between the arms. The arms have a length of $10$. \Cref{fig:hingeSetup} visualizes the shape. We compare different alignments with the ground truth parametric model: the target-specific $\GP(\mu^X, k^X)$ and our $\GP(\mathcal{P}^X[\mu^\Omega],\mathcal{P}^Xk^\Omega\mathcal{P}^{X^T})$. Both are linear models and cannot accurately model the parametric rotations.

\begin{figure}[!t]%
    \centering
    \includegraphics[width=0.5\linewidth]{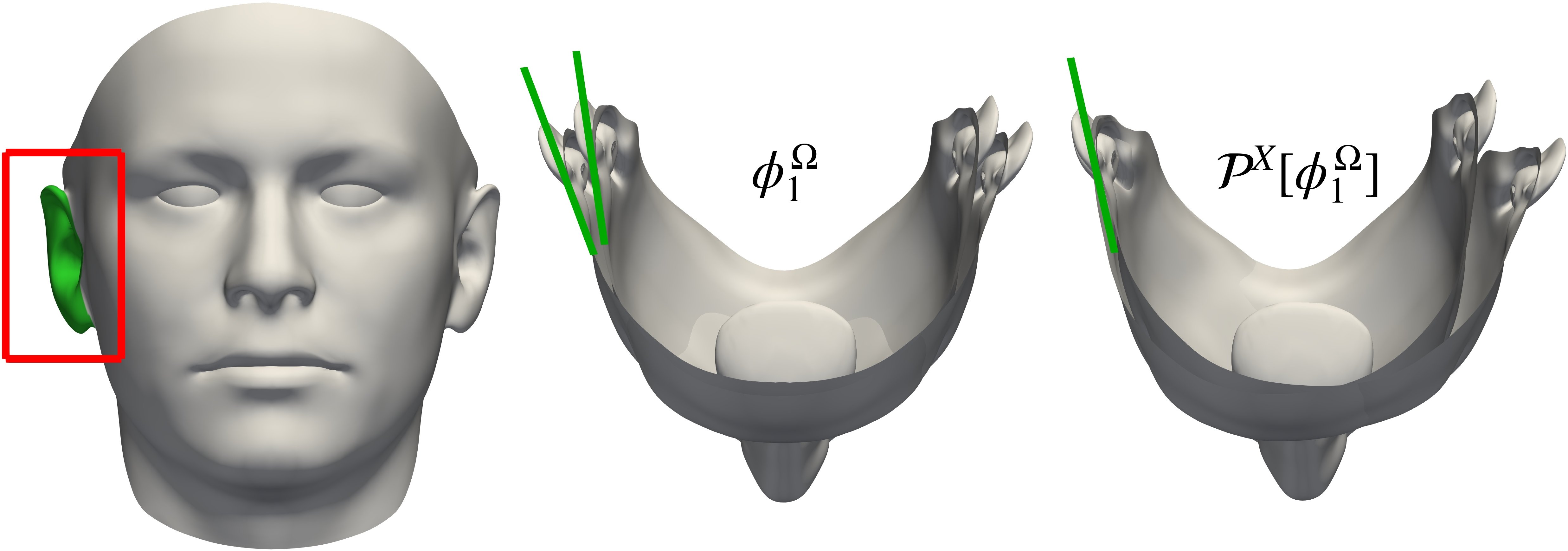}%
    \caption{Three meshes of the BFM is shown. The left mesh shows the right ear as alignment domain $X$ highlighted on the mean mesh. The middle shows $\pm3$ standard deviations of the eigenfunction $\phi_1^\Omega$ from the model agnostic basis. The green line is added to indicate the rotation of $X$. The right shows the same for our projected $\mathcal{P}^X[\phi_1^\Omega]$. The pose of the observed ear stays constant for the visualized eigenfunction.}%
    \label{fig:bfmdomainPhi}%
\end{figure}

\begin{figure}[!t]%
    \centering
    \includegraphics[width=0.5\linewidth]{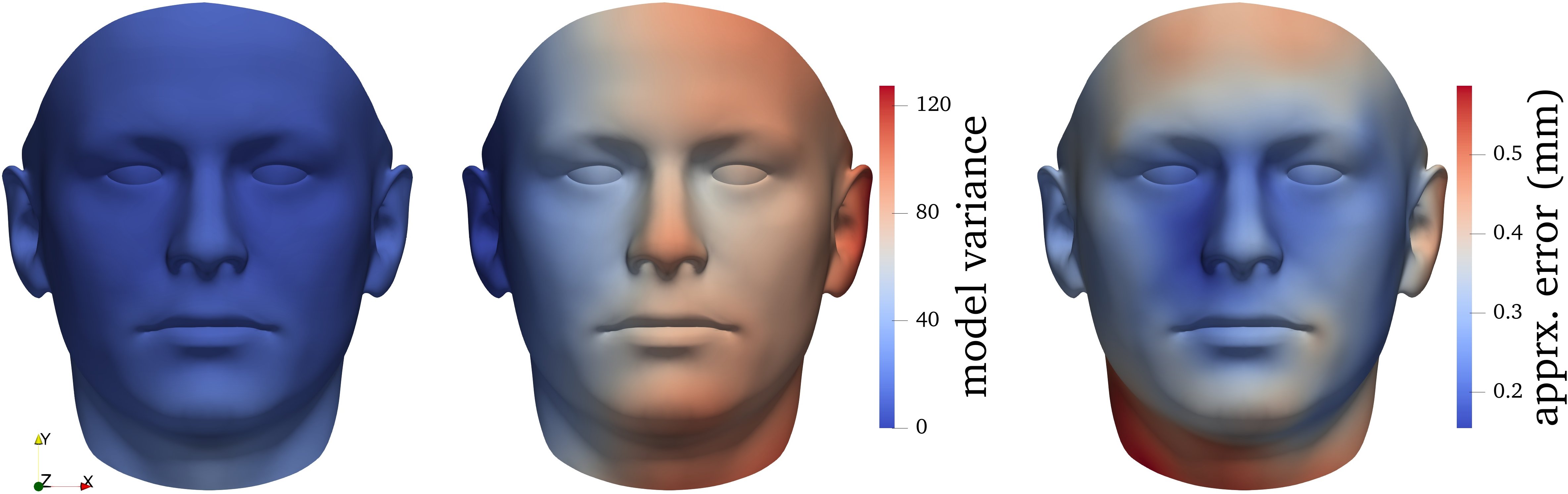}%
    \caption{The two meshes on the left show the point wise model variance of the target-agnostic and target-specific model respectively, with the alignment domain $X$ being the right ear (see \cref{fig:bfmdomainPhi}). The third mesh shows the average rotation approximation error. }%
    \label{fig:bfmvarianceError}%
\end{figure}

\Cref{fig:partialResHinge} shows the recovered parameter distribution of the three parameters: the aligned arm with domain $X$, the arm to predict $\Omega-X$ and the in between angle $\beta$. The relative error for the estimated length of the arm $X$ at $\varphi = 0.3$ is around $0.025$. The approximation quality of both models starts to be problematic at higher angles. A major reason for that is with the higher $\varphi$ values more and more shapes of the true distribution start to exhibit close or over $90\deg$ deviation to the mean $\pi$. A linear approximation of such rotations is not reasonable anymore. The projection also extends the approximation error to the predicted arm as well. The target-specific model relies on a non linear pose optimization for the arm $X$ and correctly models the remaining constant component on $X$. 

We further test the approximation accuracy for a model of a more complex shape with the BFM face model \citep{gerig2018morphable}. The training data of the shape model is aligned to a point close to the center point behind the nose. We employ our method to realign the shape model to the right ear and then measure the average error between shapes that should be equivalent when rigidly aligned to each other. \Cref{fig:bfmdomainPhi} illustrates the domain $X$ around the right ear as well as the influence of the projection on a specific eigenfunction. \Cref{fig:bfmvarianceError} visualizes the model variance of the target-agnostic and target-specific model, as well as where the approximation errors occur. The average approximation error is in expectation slightly below $0.4$ mm. For a realignment over only translation this distance is exactly $0.0$ mm as it should be.

\subsection{Different data modality}
\label{sec:scaphoids}

\begin{figure*}%
    \centering
    \subfloat{{\includegraphics[width=0.49\textwidth]{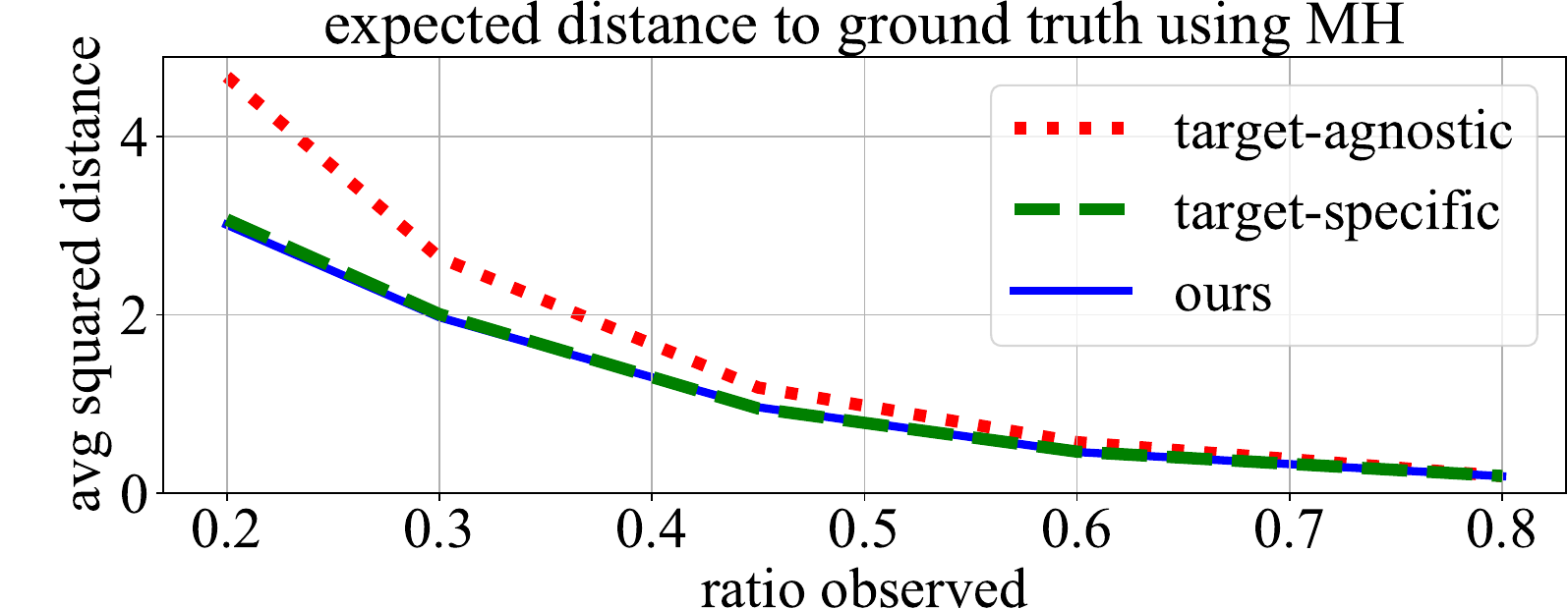} }}%
    \hfill
    \subfloat{{\includegraphics[width=0.49\textwidth]{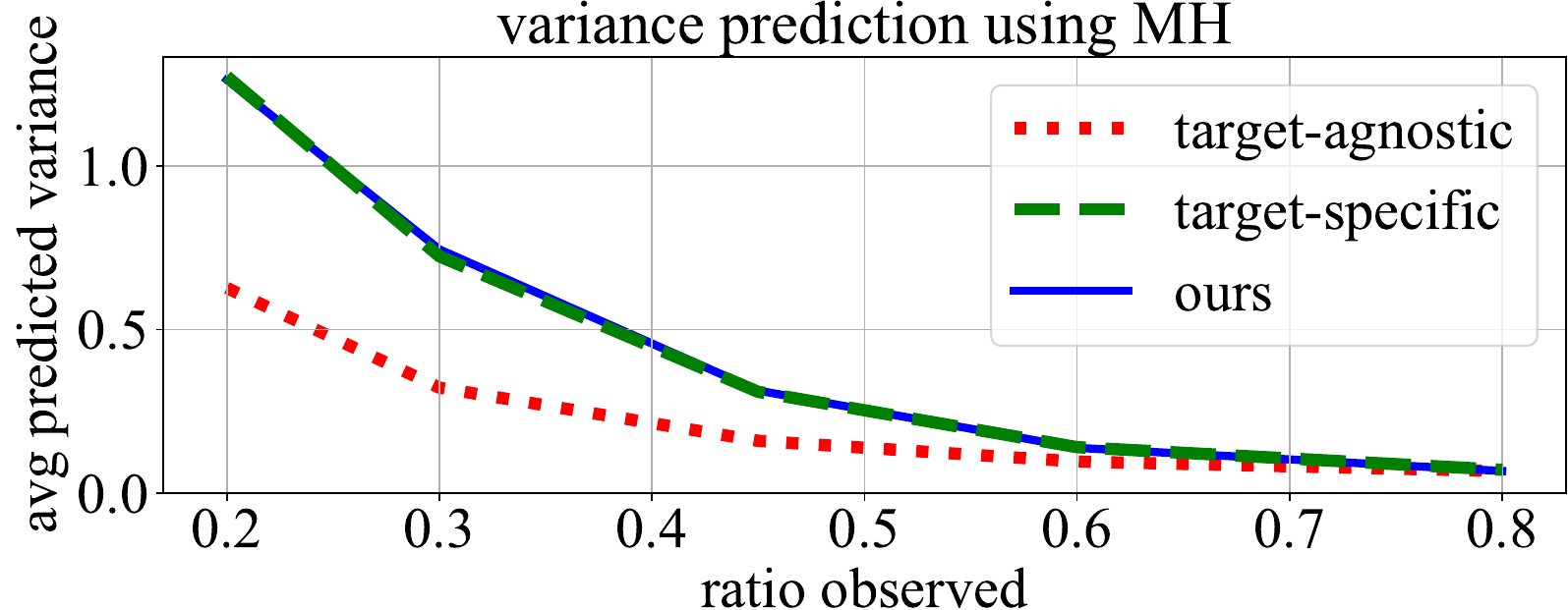} }}%
    \caption{The results using the scaphoid dataset. This is a repeat of the leave-one-out reconstruction experiment from \cref{fig:mhPost}. The left shows average distance to ground truth complete shape. The right shows the posterior uncertainty. The $x$-axis shows how much of the target is observed. The performance generalizes to a completely different shape.}%
    \label{fig:scaphoidMCMC}%
\end{figure*}

We also perform a reconstruction task with a scaphoid dataset \cite{akhbari2019predicting}, containing 120 scaphoids of which we use the 80 right-handed ones. The scaphoid represents a very different shape compared to femurs and is useful to verify the performance for shorter and rounder bones. Additionally, we simplify the problem slightly by using a less peaked likelihood in \cref{eq:fullFormulationLikelihood}. In practice this can be achieved by using a larger $\sigma$ or using fewer discretization points.

We repeat the leave-one-out reconstruction performed in \cref{sec:fullInference}. Because we simplified the likelihood, we reduced the sampling to only 15k iterations with 1k being discarded as burn-in. If more resources are used for the reconstruction by taking more samples, the target-agnostic model can overcome the ill-suited shape-pose entanglement. For example, with 80k iterations the target-agnostic model was able to reach the performance of the target-specific model which mostly achieved their peak performance with around 8k iterations. In this experiment, we report our model using the ground-truth $X$.

\Cref{fig:scaphoidMCMC} shows that the reconstruction loss is fairly similar between the target-specific model spaces, which outperform target-agnostic models. The variance prediction is also better with target-specific models. This repeats our findings on MH for another shape.

\subsection{Reconstructions using posterior shape models}
\label{sec:detailedPosteriorShapeModels}

\begin{figure*}%
    \centering
    \subfloat{{\includegraphics[width=0.49\textwidth]{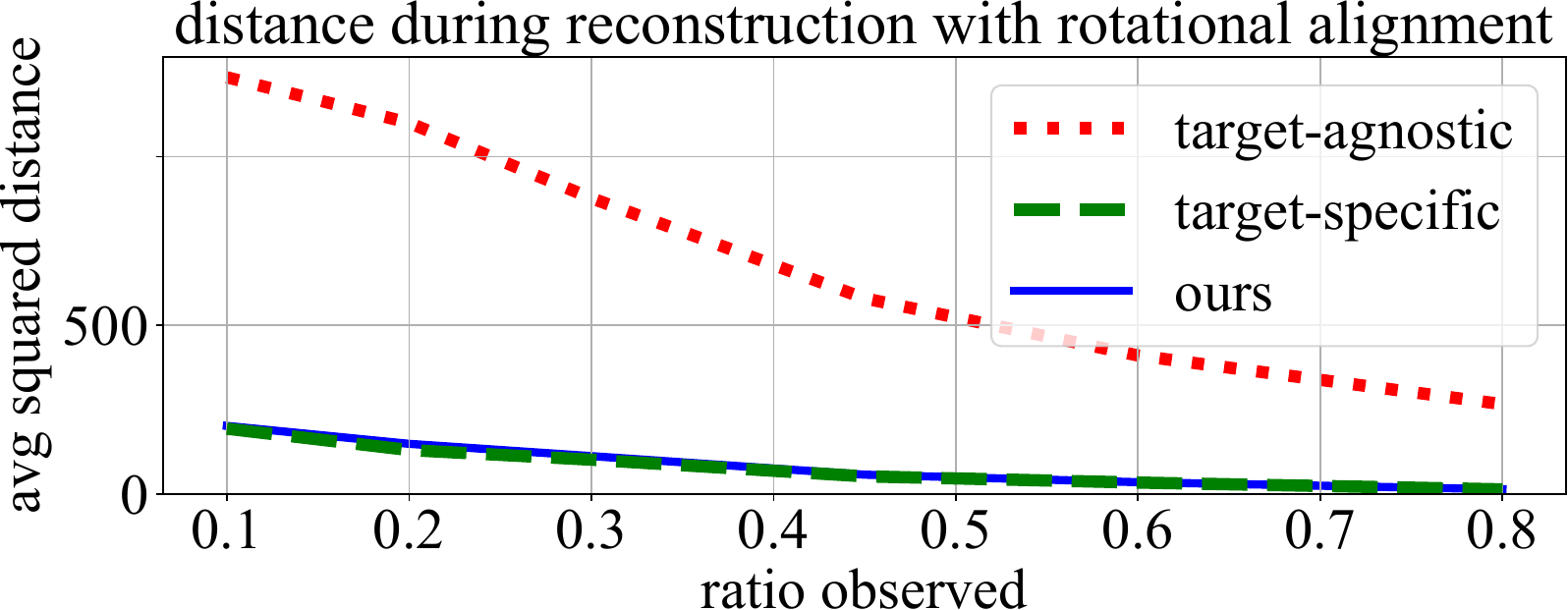} }}%
    \hfill
    \subfloat{{\includegraphics[width=0.49\textwidth]{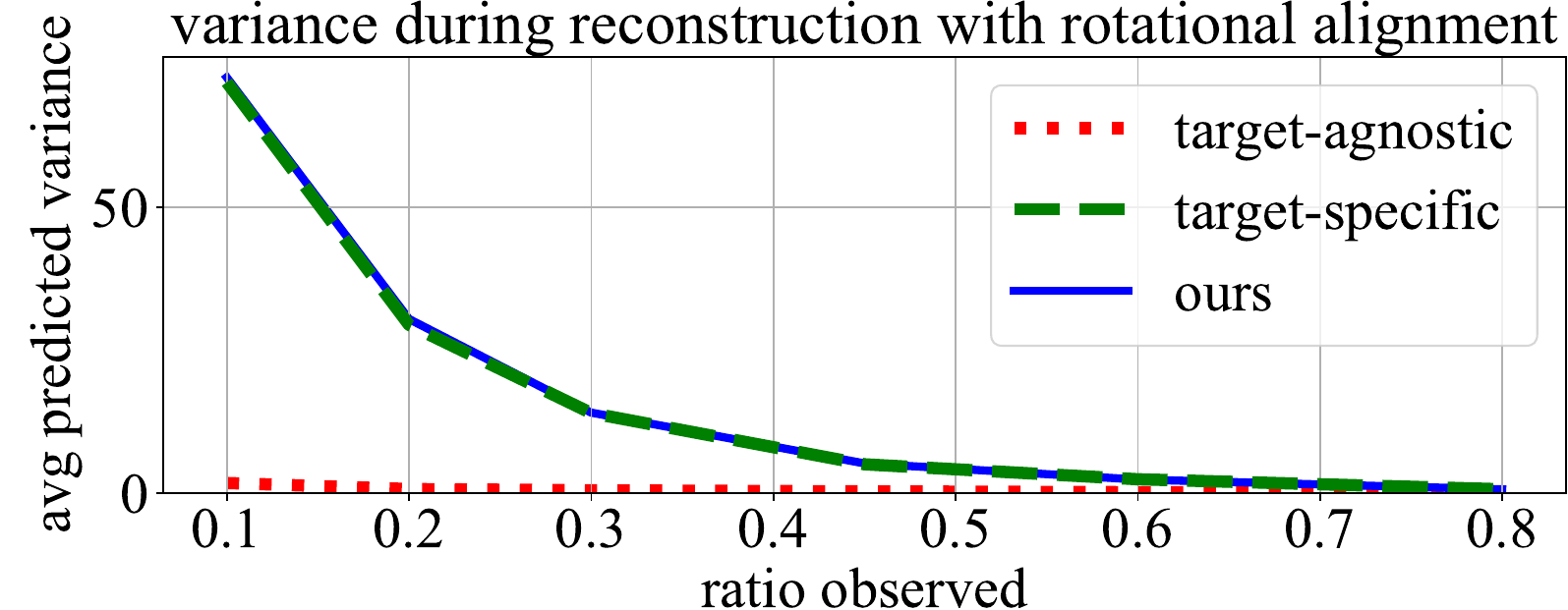} }}%
    \caption{The results of a leave-one-out femur reconstruction experiment using posterior shape models where alignment includes rotational alignment. The left shows average distance to ground truth complete shape. The right shows the posterior uncertainty. The $x$-axis shows how much of the target is observed. The benefits of using target-specific models is even greater in this setting where the pose and correspondence optimization cannot compensate the bias in the shape space.}%
    \label{fig:anaPostRot}%
\end{figure*}

It is useful to understand the influence of the alignment on the Gaussian process posterior in the context of partial shape reconstructions. Posterior shape models \citep{albrecht2013posterior} give the posterior shape parameters given a fixed pose and target. In many cases it is a reasonable first approximation of the full posterior and a tool to understand challenges for inference methods for the full shape and pose posterior.

As seen in \cref{sec:issuesReco} target-agnostic models lead to biased reconstructions. So far we only demonstrated this for in-model targets. To fill that gap we repeat the leave-one-out femur reconstruction experiment from \cref{sec:fullInference}. This time we use the ground-truth correspondence and can use analytical posteriors. We compare the target-agnostic $\GP(\mu^\Omega, k^\Omega)$, the target-specific $\GP(\mu^X, k^X)$ and our $\GP(\mathcal{P}^X[\mu^\Omega], \mathcal{P}^X k^\Omega \mathcal{P}^{X^T})$ model. For our model we use the rotation approximation from \cref{sec:alignmentRot}.

It is important to note that this experiment does \textit{not} investigate the full posterior $p(\theta | \Gamma_\tau)$ defined in \cref{eq:fullFormulation}. It only compares posterior shape models, which rely on Gaussian process regression with fixed correspondence. In other words, the pose is fixed and the optimal shape distribution is calculated based on the found correspondence.

Every observed partial target is rigidly aligned to the model following \cref{eq:gpaAlignmentOperator}. We assume a noise standard deviation of $1.0mm$ for the subsequent standard Gaussian process regression.

\Cref{fig:anaPostRot} reports the average squared distance and average predicted variance of the Gaussian process posterior. The target-agnostic model exhibits a significantly worse performance compared to the other two models. The reconstruction has a high error and almost no associated uncertainty. The target-specific and our projected model have very similar performance. This shows that the very limited rotation on $X$ is well approximated by a linear transformation. This confirms the findings of \cref{sec:issuesReco} that indicate a biased reconstruction for target-agnostic models.

Overall using target-specific models leads to a fairly competitive reconstruction loss compared to more expensive inference algorithms. This indicates a great advantage of the target-specific models, as posterior shape models are very cheap to compute. As such our projection operator allows to get target-specific posterior shape models even if only an arbitrary existing shape model is available. However, in the main paper we focus on the full pipeline that also optimizes the correspondence.

This difference in performance between the target-specific and target-agnostic models is the largest contributor to the results seen in \cref{fig:mhPost} from \cref{sec:fullInference}. The full posterior in \cref{eq:fullFormulation} can be understood as a combination of many posterior shape models for different poses and correspondences. Target-specific models require far fewer such posterior shape models to adequately represent the full posterior compared to target-agnostic models.

\section{Localizing the skull model}
\label{sec:detailedModelSkull}

The skull model is built using 46 scans of healthy shapes from an internal dataset. Following \cite{wilms2020kernelized}, we further localize the model to showcase shape models that do not use an empirical covariance function. The localization kernels are Gaussian kernels weighted by $w_i$ and using a standard deviation of $\sigma_i$:

\begin{equation}
\label{eq:skullMixture}
(w_i)_i = [0.1, 0.3, 0.4, 0.2] \quad (\sigma_i)_i = [\infty, 200, 100, 50].
\end{equation}

The $\sigma_0=\infty$ denotes the original empirical covariance function. Given the resulting covariance function $C$, we use $\mathcal{P}^\Omega C \mathcal{P}^{\Omega^T}$ as the target-agnostic model and $\mathcal{P}^X C \mathcal{P}^{X^T}$ as the target-specific model. We used the rotation extension for the projection. However, for this experiment there was no noticeable difference when including the rotation basis functions. This is to be expected as the fragment is fairly large with almost no discernible rotation in the model basis.

\section{Projection of shape models}
\label{sec:projectionShapeModels}

This Sections discusses some properties of the target-agnostic $\GP(\mu^\Omega, k^\Omega)$ and the target-specific $\GP(\mu^X, k^X)$. In this Section, we will also restrict the alignment in \cref{eq:gpaAlignmentOperator} to position alone. That means that we keep the orientation of the input data unchanged.

The model $\GP(\mu^\Omega, k^\Omega)$ and $\GP(\mu^X, k^X)$ remain a linear combination of the training data. That means that the two models still generate the same shapes when correcting for translation. As a consequence of the training data differing from each other only by a translation, the linear combination also needs to be adjusted only by a certain translation. Therefore we can summarize that there is a one to one mapping between $\GP(\mu^\Omega, k^\Omega)$ and $\GP(\mu^X, k^X)$ that requires only a translation of the shapes. This can be seen by representing the shapes as a linear combination of the training data. Let $u^\Omega \sim \GP(\mu^\Omega, k^\Omega)$ then there is a $u^X$ with the same coefficients of the linear combination. 

\begin{equation}
\label{th:linearCombination}
\begin{split}
    u^\Omega(x) &= \sum_i^n z_i u^\Omega_i(x),~~ u^X(x) = \sum_i^n z_i u^X_i(x) \\
    u^\Omega(x) &= u^X(x) - \sum_i^n z_i t_i,~~ t_i = t^X_i - t^\Omega_i.
\end{split}
\end{equation}
This is a direct consequence of the rigid alignment in \cref{eq:gpaAlignmentOperator}. Remember, for this Section we simplify the GPA as alignment over position. Indeed any shape of a model aligned on some $X$ needs only be translated to reach a shape space of another arbitrary domain $Z \subseteq \Omega$.

We can go even further than stating that when controlled for translation the probability of the shape parameters in the respective shape spaces are the same. To see this, one should remember that the probability density of the shapes are defined by the coefficients $\alpha_i$ in \cref{eq:klSum} and not the deformation fields themselves. This means that the probability density is formulated as $p^X(\boldsymbol\alpha^X)$ for some deformation field $u^X \sim \GP(\mu^X,k^X)$. \Cref{th:linearCombination} states that there is a certain translation $t$ that moves $u^X$ to some other shape space $\GP(\mu^Z, k^Z)$ with $Z \subseteq \Omega$. This translation is required as the Gaussian processes are degenerate, meaning in general $u^X$ itself is almost surely not in $\GP(\mu^Z, k^Z)$, but $u^X(x)+t$ is. More concisely, the same probability density when controlling for translation relates to $u^X$ with $\boldsymbol\alpha^X$ for $\GP(\mu^X,k^X)$ and $u^X(x) + t$ with $\boldsymbol\alpha^Z$ for $\GP(\mu^Z,k^Z)$. Indeed, they have the same density:

\begin{equation}
\label{th:pxeqpy}
p^X(\boldsymbol\alpha^X) = p^Z(\boldsymbol\alpha^Z).
\end{equation}

That the probability densities of the coefficients are the same can be understood by taking a sampling perspective. \Cref{sec:alignment} introduced $\mathcal{P}^Xk^\Omega \mathcal{P}^{X^T}$ which can be directly used to create samples

\begin{equation}
u^X(x) = \sum_{i=1}^r \alpha_i^\Omega \sqrt{\lambda_i^\Omega} \mathcal{P}^X[\phi_i^\Omega](x),
\end{equation}
where the $\alpha_i^\Omega$ are normal distributed. That the prior probability of $\boldsymbol\alpha^X$ and $\boldsymbol\alpha^Z$ are the same in their respective model can be seen from our projection $\mathcal{P}^X$. The null space of $\mathcal{P}^X$ is the component-wise constant deformation fields, what we call translations. Therefore, the projection is rank preserving. The Mahalanobis distance is constant under any linear transformation that is full rank on the shape space of $k^X$.

\end{document}